%% file: main.tex
\documentclass[10pt,twocolumn,letterpaper]{article}

\usepackage{cvpr}
\usepackage{times}
\usepackage{epsfig}
\usepackage{graphicx}
\usepackage{amsmath}
\usepackage{amssymb}
\usepackage{booktabs}

\input{math_commands.tex}

\usepackage{url}
\usepackage{graphicx}
\usepackage{amsmath}
\usepackage{amssymb}
\usepackage{csquotes}
\usepackage{makecell}
\usepackage{multirow}
\usepackage{color, colortbl}
\usepackage{xcolor}
\usepackage{subcaption}
\definecolor{Gray}{gray}{0.9}

\graphicspath{ {./images/} }

\usepackage[pagebackref=true,breaklinks=true,letterpaper=true,colorlinks,bookmarks=false]{hyperref}

\cvprfinalcopy 

\begin{document}

\title{TriGAN: Image-to-Image Translation for Multi-Source Domain Adaptation}

\author{Subhankar Roy\textsuperscript{1,2}, Aliaksandr Siarohin\textsuperscript{1}, Enver Sangineto\textsuperscript{1}, Nicu Sebe\textsuperscript{1} and Elisa Ricci\textsuperscript{1,2} \\
\textsuperscript{1}DISI, University of Trento, Italy, \textsuperscript{2}Fondazione Bruno Kessler, Trento, Italy\\
{\tt\small \{subhankar.roy, aliaksandr.siarohin, enver.sangineto, niculae.sebe, e.ricci\}@unitn.it}
}

\maketitle

\begin{abstract}
Most domain adaptation methods consider the problem of transferring knowledge to the target domain from a single source dataset. However, in practical applications, we typically have access to multiple sources. In this paper we propose the first approach for Multi-Source Domain Adaptation (MSDA) based on Generative Adversarial Networks. Our method is inspired by the observation that the appearance of a given image depends on three factors: the \emph{domain}, the \emph{style} (characterized in terms of low-level features variations) and the \emph{content}. For this reason we propose to project the image features onto a space where only the dependence from the content is kept, and then re-project this invariant representation onto the pixel space using the target domain and style. In this way, new labeled images can be generated which are used to train a final target classifier. We test our  approach using common MSDA benchmarks, showing that it outperforms state-of-the-art methods. 
\end{abstract}


\input{intro.tex}
\input{related.tex}
\input{method.tex}

\input{experiments.tex}

\section{Conclusions}

In this work we proposed TriGAN, an MSDA framework which is based on data-generation from multiple source domains using a single generator. The underlying principle of our approach to to obtain intermediate, domain and style invariant representations in order to simplify the generation process. Specifically, our generator progressively removes style and domain specific statistics from the source images and then re-projects the intermediate features onto the desired target domain and style.
We obtained state-of-the-art results on two MSDA datasets, showing the potentiality of our approach. 

{\small
\bibliographystyle{ieee_fullname}
\bibliography{cvpr2020_conference}
}

\clearpage
\input{arxiv-supp.tex}

\end{document}

%% file: math_commands.tex
\usepackage{amsmath,amsfonts,bm}

\def\eqref#1{equation~\ref{#1}}

\def\1{\bm{1}}

\DeclareMathAlphabet{\mathsfit}{\encodingdefault}{\sfdefault}{m}{sl}
\SetMathAlphabet{\mathsfit}{bold}{\encodingdefault}{\sfdefault}{bx}{n}



%% file: intro.tex
\section{Introduction}
\vspace{-0.2cm}
\label{Introduction}

A well known problem in computer vision is the need to adapt a classifier trained on a given {\em source} domain in order to work on a different,
{\em target}  domain.
Since the two domains typically have different marginal feature distributions, the adaptation process needs to 
reduce the corresponding
{\em domain shift} \cite{torralba2011unbiased}. In many practical scenarios, the target data are not annotated and  
Unsupervised Domain Adaptation (UDA) methods are required.
 
While most previous adaptation approaches consider a single source domain, in real world applications we may have access to multiple datasets. In this case, Multi-Source Domain Adaptation (MSDA) methods \cite{yao2010boosting,mansour2009domain,xu2018deep,peng2018moment}  may be adopted, in which more than one source dataset is considered in order to make the  adaptation process more robust. However, despite more data can be used, MSDA is challenging as multiple domain-shift problems need to be simultaneously and coherently solved.

In this paper we deal with  (unsupervised)  MSDA  using a data-augmentation approach based on
 a Generative Adversarial Network (GAN) \cite{goodfellow2014generative}. Specifically, we  generate artificial target samples 
by ``translating'' images from all the source domains into target-like images. Then the synthetically generated images are used
for training the target classifier. While this strategy has been recently adopted in the single-source UDA scenario \cite{russo17sbadagan,hoffman2017cycada,liu2016coupled,murez2018image,sankaranarayanan2018generate}, 
we are the first to show how it can be effectively used in a MSDA setting.
In more detail, our goal is to build and train a ``universal'' translator which can transform an image from an input domain to a target domain. The translator network is ``universal'' because the number of parameters which need to be optimized should scale linearly with the number of domains. We achieve this goal using domain-invariant intermediate features, computed  by the {\em encoder} part of our generator, and then projecting these features onto the domain-specific target distribution using the {\em decoder}.

To make this image translation effective,  we assume that the appearance of an image depends on three factors: the {\em content}, the {\em domain} and the {\em style}. The \emph{domain} models properties that are shared by the elements of a dataset but which may not be shared by other datasets.
On the other hand, the  \emph{style} factor represents properties which are shared among different {\em local} parts of {\em a single image} and describes low-level features which concern a specific image (e.g., the color or the texture).
The \emph{content} is what we want to keep unchanged during the translation process: typically, it is the foreground object shape which is described by the image labels associated with the source data samples.
Our encoder obtains the intermediate representations in a two-step process: we first generate style-invariant representations and then we compute the  domain-invariant representations. Symmetrically, the decoder transforms the intermediate representations first projecting these features onto a domain-specific distribution and then onto a style-specific distribution.
In order to modify the underlying distribution of a set of features, 
inspired by \cite{roy2019unsupervised}, in the encoder we use \textit{whitening} layers which progressively align the style-and-domain feature distributions. Then, in the decoder, we project the intermediate invariant representation
onto a new domain-and-style specific distribution with \emph{Whitening and Coloring} ($WC$) \cite{2018arXiv180600420S} batch transformations, according to the target data.

A ``universal'' translator similar in spirit to our proposed generator is StarGAN \cite{choi2018stargan} (proposed in a non UDA  task).  However, in StarGAN the domain information is represented by a one-hot vector concatenated with the input image. 
When we use StarGAN in our MSDA scenario, the synthesized images are much less effective for training the target classifier, and this emiprically shows that our batch-based transformation of the image distribution is more effective for our translation task.

    

\noindent \textbf{Contributions.} Our main contributions can be summarized as follows.
(i) We propose the first generative  MSDA method. We call our approach TriGAN because it is based on three different factors of the images:
 the style, the domain and the content. (ii) The proposed image translation process is based on style and domain specific statistics which are first removed from and then added to the source images by means of modified  $WC$ layers. Specifically, we use the following feature transformations (associated with a corresponding layer type): Instance Whitening Transform ($IWT$), Domain Whitening Transform ($DWT$) \cite{roy2019unsupervised}, conditional Domain Whitening Transform ($cDWT$) and Adaptive Instance Whitening Transform ($AdaIWT$). $IWT$ and $AdaIWT$ are novel layers introduced in this paper.  
(iii) We test our method on two MSDA datasets, Digits-Five \cite{xu2018deep} and Office-Caltech10 \cite{gong2012geodesic}, outperforming state-of-the-art methods. 

%% file: related.tex
\begin{figure*}[t]
    \centering
    \includegraphics[width=1\textwidth]{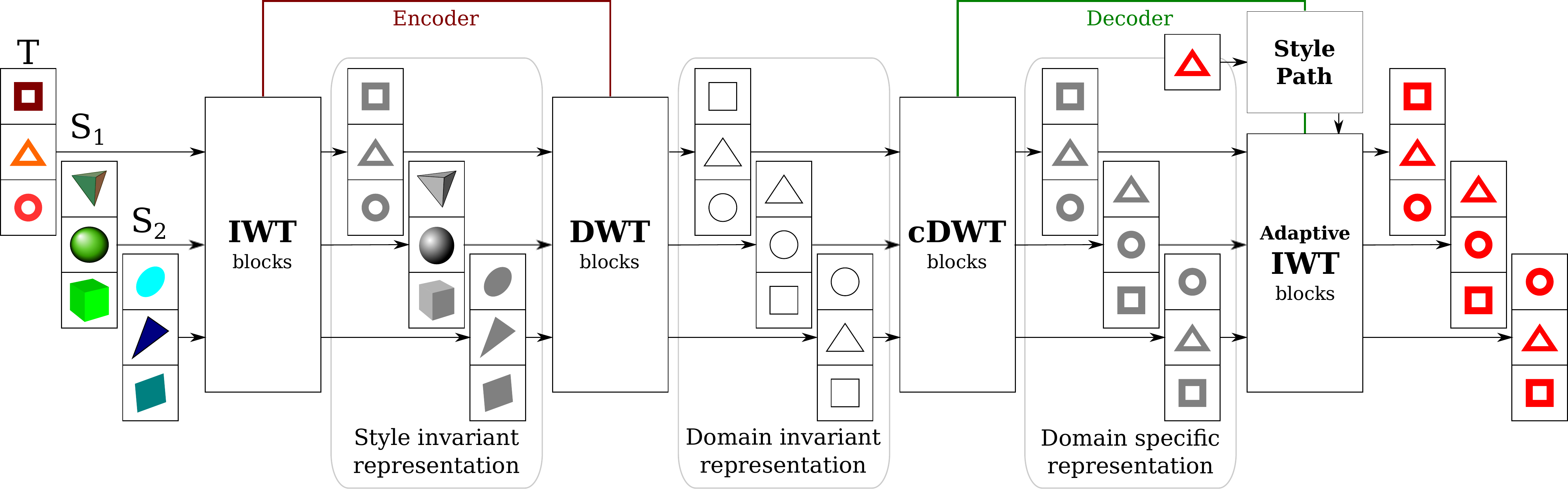}
    \caption{An overview of the TriGAN generator. We schematically show 3 domains $\{T, S_1, S_2\}$ - objects with \textit{holes}, \textit{3D objects} and \textit{skewered} objects, respectively. The content is represented by the object's shape  - square, circle or triangle. The style is represented by the color: each image input to $\mathcal{G}$ has a different color and each domain has it own set of styles. First, the encoder $\mathcal{E}$ creates a style-invariant representation using IWT blocks.   DWT blocks are then used to obtain a domain-invariant representation. Symmetrically, the decoder $\mathcal{D}$ brings back domain-specific information with cDWT blocks (for simplicity we show only a single output domain, $T$). Finally, we apply  a reference style. The reference style is extracted using the style path and it is applied using the Adaptive IWT blocks.
    }
    
    \label{fig:overall_pipenline}
\end{figure*}

\vspace{-0.2cm}
\section{Related Work}
\vspace{-0.2cm}
\label{Related}

In this section we review the previous approaches on UDA, considering both single source and multi-source methods. Since the proposed generator is also related to deep models used for image-to-image translation, we also analyse related work on this topic.

\noindent \textbf{Single-source UDA}. Single-source UDA approaches assume a single labeled source domain and can be broadly classified under three main categories, depending upon the strategy adopted to cope with the domain-shift problem. The first category uses  first and second order statistics to model the source and the target feature distributions. For instance, \cite{long2015learning, long2016deep, venkateswara2017deep, tzeng2014deep} minimize the Maximum Mean Discrepancy, \textit{i.e.} the distance between the mean of feature distributions between the two domains. On the other hand, \cite{sun2016deep, morerio2017minimal, peng2018synthetic} achieve domain invariance by aligning the second-order statistics through correlation alignment.
Differently, \cite{carlucci2017autodial, li2016revisiting, mancini2018boosting} reduce the domain shift by domain alignment layers derived from batch normalization (BN) \cite{ioffe2015batch}. This idea has been recently extended in \cite{roy2019unsupervised}, where grouped-feature whitening (DWT) is used instead of feature standardization as in $BN$. In  our proposed encoder we also use the DWT layers,  which we adapt to work in a generative network. In addition, we also propose other style and domain dependent batch-based normalizations (i.e., $IWT$, $cDWT$ and $AdaIWT$). 

The second category of methods computes domain-agnostic representations by means of  an adversarial learning-based approach. For instance, discriminative domain-invariant representations are constructed through a gradient reversal layer in \cite{ganin2014unsupervised}.  Similarly, the approach in \cite{Hoffman:Adda:CVPR17} uses  a domain confusion loss to promote the alignment between the source and the target domain. 

The third category of methods uses adversarial learning in a generative framework (i.e., GANs \cite{goodfellow2014generative}) to reconstruct artificial source and/or target images and perform domain adaptation. Notable approaches are SBADA-GAN \cite{russo17sbadagan}, CyCADA \cite{hoffman2017cycada}, CoGAN \cite{liu2016coupled}, I2I Adapt \cite{murez2018image} and Generate To Adapt (GTA) \cite{sankaranarayanan2018generate}. 
While these generative methods have been shown to be very successful in UDA, none of them deals with a multi-source setting. 
Note that trivially extending these approaches to an MSDA scenario involves training $N$ different generators, being $N$ the number of source domains.
In contrast, in our universal translator only a subset of parameters  grow linearly with the number of domains (Sec.~\ref{Decoder}), while the others are shared over all the domains. Moreover, since we train our generator using
  $(N+1)^2$  translation directions, we can largely increase the number of training sample-domain pairs effectively used
 (Sec.~\ref{sec:net_train}).

\noindent \textbf{Multi-source UDA}. In \cite{yao2010boosting},  multiple-source knowledge transfer is obtained by borrowing knowledge from the target \textit{k} nearest-neighbour sources.
Similarly, a distribution-weighted combining rule is proposed in \cite{mansour2009domain} to construct a target hypothesis as a weighted combination of source hypotheses. Recently, Deep Cocktail Network (DCTN) \cite{xu2018deep} uses the distribution-weighted combining rule in an adversarial setting. A Moment Matching Network ($\textrm{M}^{3}\textrm{SDA}$) is introduced in \cite{peng2018moment} to reduce the discrepancy between the multiple source and the target domains. Differently from these methods which operate in a discriminative setting, our method relies on a deep generative approach for MSDA.

\noindent \textbf{Image-to-image Translation}. Image-to-image translation approaches, i.e.\ those methods which learn how to transform an image from one domain to another, possibly keeping its semantics, are the basis of our method. In \cite{isola2017image} the pix2pix network translates images under the assumption that paired images in the two domains are available at training time. In contrast,  CycleGAN \cite{zhu2017unpaired} can learn to translate images using unpaired training samples. Note that, by design, these methods  work with two domains. ComboGAN \cite{anoosheh2018combogan} partially alleviates this issue by using $N$ generators for translations among $N$ domains. Our work is also related to StarGAN \cite{choi2018stargan} which handles unpaired image translation amongst \textit{N} domains (\textit{N} $\geq$ 2) through a single generator. However, StarGAN achieves image translation without explicitly forcing the image representations to be domain invariant, and this may lead to a significant reduction of the network representation power as the number of domains increases. On the other hand, our goal is to obtain an explicit, intermediate image representation  which is style-and-domain independent. We use \textit{IWT} and \textit{DWT} to achieve this. We also show that this invariant representation can simplify the re-projection process onto a desired style and target domain. This is achieved through $AdaIWT$ and $cDWT$ which results into very realistic translations amongst domains. Very recently, a whitening and colouring based image-to-image translation method was proposed in \cite{cho2019image}, where the whitening operation is {\em weight-based}: the transformation is embedded into the network weights.  Specifically,  whitening  is approximated by enforcing the convariance matrix, computed using the intermediate features, to be equal to the identity matrix. Conversely, our whitening transformation is {\em data dependent} (i.e., it depends on the specific batch statistics, Sec. \ref{Preliminaries}) and uses the Cholesky decomposition \cite{dereniowski2003cholesky} to compute the whitening matrices of the input samples in a closed form, thereby eliminating the need of additional ad-hoc losses.


%% file: method.tex
\vspace{-0.2cm}
\section{Style-and-Domain based Image Translation}
\label{Method}
\vspace{-0.2cm}
In this section we describe the proposed approach for MSDA. We first provide an overview of our method and we introduce the notation adopted throughout the paper (Sec. \ref{Overview}). Then we describe the TriGAN architecture   (Sec. \ref{Generator}) and our training procedure (Sec.\ref{GeneratorTraining}).

\subsection{Notation and Overview}
\label{Overview}

In the MSDA scenario we have access to $N$ labeled source datasets $\{ S_{j} \}_{j=1}^N$, where 
$S_j = \{(\mathbf{x}_k, y_k)\}_{k=1}^{n_j}$,
and a target unlabeled dataset $T = \{\mathbf{x}_k \}_{k=1}^{n_t}$.
All the datasets (target included) share the same categories and each of them is associated to a domain 
$\mathbf{D}_1^s, ...,\mathbf{D}_N^s, \mathbf{D}^t$, respectively. 
 Our final goal is to build a classifier for the target domain $\mathbf{D}_t$ exploiting the data in $\{S_{j} \}_{j=1}^N \cup T$.
 
Our method is based on two separate training stages. We initially train a generator $\mathcal{G}$ which learns how to change the appearance of a real input image in order to adhere to a desired  domain and style. Importantly, our $\mathcal{G}$ learns mappings between every possible pair of image domains.
Learning $(N+1)^2$ translations makes it possible to exploit much more supervisory information with respect to a plain strategy in which $N$ different source-to-target generators are trained (Sec.~\ref{sec:net_train}).
Once $\mathcal{G}$ is trained, in the second stage we use it to generate target data having the same content  of the source data, thus creating a new, {\em labeled}, target dataset, which is finally used  to train a target classifier $\mathcal{C}$. 
However, 
in training $\mathcal{G}$ (first stage),
we do not use 
 class labels 
 and $T$ is treated in the same way as the other  datasets.
 
As mentioned in Sec.~\ref{Introduction}, $\mathcal{G}$ is composed of an encoder $\mathcal{E}$ and a decoder $\mathcal{D}$ (Fig.~\ref{fig:overall_pipenline}).
The role of  $\mathcal{E}$ is to ``whiten'', \textit{i.e.}, to remove, both domain-specific and style-specific aspects of the input image features in order to obtain domain and style invariant representations. Symmetrically, 
$\mathcal{D}$  ``colors'' the domain-and-style invariant features generated  by $\mathcal{E}$,
by progressively projecting  these intermediate representations  onto a  domain-and-style specific space.

 In the first training stage, $\mathcal{G}$ takes as input a batch of images $B = \{ \mathbf{x}_1, ..., \mathbf{x}_m \}$ with corresponding {\em domain} labels  $L = \{ l_1, ..., l_m \}$, where $\mathbf{x}_i$ belongs to the domain $\mathbf{D}_{l_i}$ and $l_i \in [1, N+1]$. Moreover, $\mathcal{G}$ takes as input a batch of output domain labels $L^O = \{ l_1^O, ..., l_m^O \}$, and a batch of reference style images $B^O = \{ \mathbf{x}^O_1, ..., \mathbf{x}^O_m \}$, such that $\mathbf{x}^O_i$ has domain label $l_i^O$.
 For a given $\mathbf{x}_i \in B$, the task of $\mathcal{G}$ is to 
  transform $\mathbf{x}_i$ into $\hat{\mathbf{x}}_i$ such that: (1) $\mathbf{x}_i$ and $\hat{\mathbf{x}}_i$ share the same content 
 but (2) $\hat{\mathbf{x}}_i$ belongs to domain $\mathbf{D}_{l_i^O}$ and has the same  style of   $\mathbf{x}^O_i$.

\subsection{TriGAN Architecture}
\label{Generator}
The TriGAN architecture is composed of a generator network $\mathcal{G}$ and a discriminator network $\mathcal{D}_{\mathcal{P}}$. As  above mentioned, $\mathcal{G}$ comprises an encoder $\mathcal{E}$ and decoder $\mathcal{D}$, which we describe in (Sec.~\ref{Encoder}-\ref{Decoder}). The discriminator $\mathcal{D}_{\mathcal{P}}$  is based on the Projection Discriminator (\cite{miyato2018cgans}). Before describing the details of $\mathcal{G}$, we briefly review the $WC$ transform (\cite{2018arXiv180600420S}) (Sec.~\ref{Preliminaries}) which is used as the basic operation in our proposed batch-based feature transformations.

\subsubsection{Preliminaries: Whitening $\&$ Coloring Transform}
\label{Preliminaries}

Let $F(\mathbf{x}) \in \mathbb{R}^{h \times w \times d}$ be the tensor representing the activation values of the convolutional feature maps in a given layer corresponding to the input image $\mathbf{x}$, with $d$ channels and $h \times w$ spatial locations. We  treat each spatial location as a $d$-dimensional vector, in this way each image $\mathbf{x}_i$ contains a set of vectors $X_i = \{ \mathbf{v}_1, ..., \mathbf{v}_{h \times w} \}$. With a slight abuse of the notation, we use  $B = \displaystyle \mathop{\cup}_{i=1}^{m}{X_i}=\{ \mathbf{v}_1, ..., \mathbf{v}_{h \times w \times m} \}$, which includes all the spatial locations in all the images in a batch. The $WC$ transform is a multivariate extension of the per-dimension normalization and shift-scaling  
transform ($BN$)
proposed in (\cite{ioffe2015batch})
and widely adopted in both generative and discriminative networks. $WC$ can be described by:
\vspace{-2mm}
\begin{equation}
    \label{eq.WC-coloring-fin}
    WC(\mathbf{v}_j; B, \boldsymbol{\beta}, \boldsymbol{\Gamma})=Coloring(\bar{\mathbf{v}}_j; \boldsymbol{\beta}, \boldsymbol{\Gamma}) = \boldsymbol{\Gamma}   \bar{\mathbf{v}}_j + \boldsymbol{\beta}  
\end{equation}
where:
\vspace{-2mm}
\begin{equation}
\label{eq.WC-whitening-fin}
\bar{\mathbf{v}}_j=Whitening(\mathbf{v}_j; B) = \boldsymbol{W}_B (\mathbf{v}_j - \boldsymbol{\mu}_B). 
\end{equation}

In Eq.~\ref{eq.WC-whitening-fin},  $\boldsymbol{\mu}_B$ is the centroid of the elements in $B$, while $\boldsymbol{W}_B$ is such that: $\boldsymbol{W}_B^\top \boldsymbol{W}_B = \boldsymbol{\Sigma}_B^{-1}$, where $\boldsymbol{\Sigma}_B$ is the covariance matrix computed using $B$. 
The result of  applying Eq.~\ref{eq.WC-whitening-fin} to the elements of $B$, is a  set of {\em whitened} features  $\bar{B} = \{ \bar{\mathbf{v}}_1, ..., \bar{\mathbf{v}}_{h \times w \times m} \}$,
which
lie in a  {spherical distribution} (\textit{i.e.}, with a covariance matrix equal to the identity matrix).  
On the other hand, Eq.~\ref{eq.WC-coloring-fin} performs a {\em coloring} transform, \textit{i.e.} projects 
 the elements in $\bar{B}$ onto a {learned} multivariate Gaussian distribution.
 While  $\boldsymbol{\mu}_B$ and $\boldsymbol{W}_B$ are computed using the elements in $B$ (they are data-dependent),
 Eq.~\ref{eq.WC-coloring-fin} 
 depends on the  $d$ dimensional learned parameter vector
$\boldsymbol{\beta}$ and the $d \times d$ dimensional learned parameter matrix $\boldsymbol{\Gamma}$. Eq.~\ref{eq.WC-coloring-fin} is a linear operation and can be  simply implemented using a convolutional layer with kernel size $1 \times 1$. 

In this paper we use the WC transform in our encoder $\mathcal{E}$ and decoder $\mathcal{D}$, in order to
first obtain a style-and-domain invariant representation for each $\mathbf{x}_i \in B$, and then transform this representation 
accordingly to the desired output domain $\mathbf{D}_{l_{i}^{O}}$ and style image sample $\mathbf{x}^O_i$. The next sub-sections show the details of the proposed architecture.

\subsubsection{Encoder}
\label{Encoder}

The encoder $\mathcal{E}$ is composed of a sequence of standard $Convolution_{k \times k}$ - $Normalization$ - $ReLU$ - $Average Pooling$ blocks and some $ResBlocks$ (more details in the Supplementary Material), in which we replace the common $BN$ layers (\cite{ioffe2015batch})
with our proposed normalization modules, which are detailed below.

{\bf Obtaining Style Invariant Representations.} 
In the first two
blocks of $\mathcal{E}$ we whiten first and second-order statistics of the low-level features of each $X_i \subseteq B$, which are mainly responsible for the \textit{style} of an image (\cite{gatys2016image}). To do so, we propose the {\em Instance Whitening Transform} ($IWT$), where the term {\em instance} is inspired by Instance Normalization ($IN$) (\cite{DBLP:journals/corr/UlyanovVL16}) and highlights that the proposed  transform is applied to a set of features extracted from a single image $\mathbf{x}_i$. For each $ \mathbf{v}_j \in X_i$, 
$IWT(\mathbf{v}_j)$ is defined as:

\vspace{-2mm}
\begin{equation}
\label{IW}
IWT(\mathbf{v}_j)= WC(\mathbf{v}_j; X_i, \boldsymbol{\beta}, \boldsymbol{\Gamma}).
\end{equation}

Note that in
 Eq.~\ref{IW} we use $X_i$ as the batch, where $X_i$ contains only feautures of a specific image $\mathbf{x}_i$ (Sec. \ref{Preliminaries}).
 Moreover, each $\mathbf{v}_j \in X_i$ is extracted from the first two convolutional layers of  $\mathcal{E}$, thus  $\mathbf{v}_j$ has a small receptive field.
 This  implies that  whitening is performed using an {\em image-specific} feature centroid
$\boldsymbol{\mu}_{X_i}$
and covariance matrix $\boldsymbol{\Sigma}_{X_i}$,
which represent the first and second-order statistics of the low-level features of  $\mathbf{x}_i$.
On the other hand, coloring is based on the parameters 
$\boldsymbol{\beta}$ and $\boldsymbol{\Gamma}$, which {\em do not depend} on $\mathbf{x}_i$ or $l_i$. The coloring operation is the analogous of the shift-scaling per-dimension transform computed in $BN$ just after feature standardization (\cite{ioffe2015batch}) and is necessary to avoid decreasing the network representation capacity (\cite{2018arXiv180600420S}).

{\bf Obtaining Domain Invariant Representations.}
In the subsequent blocks of $\mathcal{E}$ we whiten first and second-order statistics which are {\em domain specific}.
For this operation we adopt the {\em Domain Whitening Transform} ($DWT$) proposed in (\cite{roy2019unsupervised}).
Specifically,
for each $X_i \subseteq B$, let $l_i$ be its domain label (see Sec.~\ref{Overview}) and let $B_{l_i} \subseteq B$ be the subset of feature which have been extracted from {\em all} those images in $B$ which share {\em the same  domain label}. 
Then, for each $\mathbf{v}_j \in B_{l_i}$:

\vspace{-2mm}
\begin{equation}
\label{DW}
DWT(\mathbf{v}_j)= WC(\mathbf{v}_j; B_{l_i}, \boldsymbol{\beta}, \boldsymbol{\Gamma}).
\end{equation}

Similarly to Eq.~\ref{IW}, Eq.~\ref{DW} performs whitening using a subset of the current feature batch. Specifically,
all the features in $B$ are partitioned depending on the domain label of the image they have been extracted from, so obtaining {$B_1, B_2, ...$, etc}, where all the features in $B_l$ belongs to the images of the domain $\mathbf{D}_{l}$.
Then, $B_l$ is used to compute domain-dependent 
first and second order statistics
($\boldsymbol{\mu}_{B_l}, \boldsymbol{\Sigma}_{B_l}$). These statistics are used 
to project each $\mathbf{v}_j \in B_l$ onto a domain-invariant spherical distribution.
A similar idea was recently proposed in (\cite{roy2019unsupervised}) in a discriminative network for single-source UDA. 
However, differently from (\cite{roy2019unsupervised}), we also use coloring by re-projecting the whitened features onto a new space
governed by a learned multivariate distribution. This is done using the (layer-specific) parameters $\boldsymbol{\beta}$ and $\boldsymbol{\Gamma}$ which do not depend on $l_i$.

\subsubsection{Decoder}
\label{Decoder}

Our decoder $\mathcal{D}$ is functionally and 
structurally symmetric with respect to $\mathcal{E}$: it takes as input the domain and style invariant 
features computed by $\mathcal{E}$ and projects these features onto the desired  domain $\mathbf{D}_{l_{i}^{O}}$ with the style extracted from the reference image $\mathbf{x}_i^O$.

Similarly to $\mathcal{E}$, $\mathcal{D}$ is a sequence of $ResBlocks$ and a few  $Upsampling$ - $Normalization$ - $ReLU$ - $Convolution_{k \times k}$ blocks (more details in the Supplementary Material).
Similarly to Sec.~\ref{Encoder}, in the $Normalization$ layers we replace $BN$ with our proposed feature normalization approaches, which are detailed below.

{\bf Projecting Features onto a Domain-specific Distribution.} 
Apart from the last two 
blocks of $\mathcal{D}$ (see below), all the other blocks are dedicated to project the current set of features onto a domain-specific subspace.
This subspace is learned from data using domain-specific coloring parameters 
$(\boldsymbol{\beta}_{l}, \boldsymbol{\Gamma}_{l})$, where $l$ is the label of the corresponding domain. To this purpose we introduce the
{\em conditional Domain Whitening Transform} ($cDWT$),
where the term ``conditional'' specifies that the coloring step is conditioned on the domain label $l$.
In more detail:
Similarly to Eq.~\ref{DW}, we first partition $B$ into 
$B_1, B_2, ...$, etc. 
However, the membership of $\mathbf{v}_j \in B$ to
 $B_{l}$ is decided taking into account the {\em desired output} domain label $l^O_i$ for each image rather than its original domain as in case of  Eq.~\ref{DW}.
 Specifically, if $\mathbf{v}_j \in X_i$ and the output domain label of $X_i$ is $l^O_i$,
 then $\mathbf{v}_j$ is included in  $B_{l^O_i}$.
Once $B$ has been partitioned, we define $cDWT$ as follows:

\vspace{-2mm}
\begin{equation}
\label{DWC}
cDWT(\mathbf{v}_j)= WC(\mathbf{v}_j; B_{l^O_i}, \boldsymbol{\beta}_{l^O_i}, \boldsymbol{\Gamma}_{l^O_i}).
\end{equation}

Note that, after whitening, and differently from  Eq.~\ref{DW},
 coloring in  Eq.~\ref{DWC} is performed using {\em domain-specific} parameters $(\boldsymbol{\beta}_{l_i^O}, \boldsymbol{\Gamma}_{l_i^O})$.

{\bf Applying a Specific Style.} In order to apply a specific style to   $\mathbf{x}_i$, we first extract the output style from the reference image $\mathbf{x}_i^O$ 
associated with $\mathbf{x}_i$ (Sec.~\ref{Overview}). This is done using
 the \textit{Style Path} (see Fig.~\ref{fig:overall_pipenline}), which consists of two $Convolution_{k \times k}$ - $IWT$ - $ReLU$ - $Average Pooling$ blocks (which share the parameters with the first two layers of the encoder) and a MultiLayer Perceptron (MLP) $\mathcal{F}$. Following (\cite{gatys2016image}) we represent a style using the first and the second order statistics ${\boldsymbol{\mu}}_{X_i^O}, \boldsymbol{W}_{X_i^O}^{-1}$, which are extracted using the $IWT$ blocks (Sec.~\ref{Encoder}). Then we use $\mathcal{F}$ to adapt these statistics to the domain-specific representation obtained as the output of the previous step. In fact, in principle, for each $\mathbf{v}_j \in X_i^O$,
the $Whitening()$ operation inside the $IWT$ transform could be ``inverted'' using:
\vspace{-0.2cm}
\begin{equation}
\label{IWC-wrong}
 Coloring(\mathbf{v}_j;  \boldsymbol{\mu}_{X_i^O}, \boldsymbol{W}_{X_i^O}^{-1}).
\end{equation}

\noindent
Indeed,  the coloring operation (Eq.~\ref{eq.WC-coloring-fin}) is the inverse of whitening (Eq.~\ref{eq.WC-whitening-fin}).
However, the elements of $X_i$ now lie in a feature space different  from the output space of Eq.~\ref{IW}, thus the transformation defined by \textit{Style Path} needs to be adapted.
For this reason, we use a MLP ($\mathcal{F}$)  which implements this adaptation:
\vspace{-0.2cm}
\begin{equation}
\label{f-MLP}
[\boldsymbol{\beta}_{i} \Vert \boldsymbol{\Gamma}_{i} ] = \mathcal{F}([\boldsymbol{\mu}_{X_i^O}  \Vert \boldsymbol{W}_{X_i^O}^{-1}]).
\end{equation}

Note that, in Eq.~\ref{f-MLP}, $[\boldsymbol{\mu}_{X_i^O}  \Vert \boldsymbol{W}_{X_i^O}^{-1}]$ is the (concatenated) input and $[\boldsymbol{\beta}_i  \Vert  \boldsymbol{\Gamma}_i ]$ is the MLP output, one  input-output pair per image $\mathbf{x}_i^O$. 

Once $(\boldsymbol{\beta}_i, \boldsymbol{\Gamma}_i )$ have been generated,
we  use them as the coloring parameters of our  {\em Adaptive IWT} ($AdaIWT$):
\vspace{-0.15cm}
\begin{equation}
\label{IWC}
AdaIWT(\mathbf{v}_j)= WC(\mathbf{v}_j; X_i^O, \boldsymbol{\beta}_i, \boldsymbol{\Gamma}_i).
\end{equation}

\noindent
Eq.~\ref{IWC} imposes style-specific first and second order statistics to the features of the last blocks of $\mathcal{D}$ in order to mimic the style of $\mathbf{x}_i^O$.

\subsection{Network Training}
\label{sec:net_train}
\textbf{GAN Training.}
\label{GeneratorTraining}
For the sake of clarity, in the rest of the paper we use a simplified notation for $\mathcal{G}$, in which $\mathcal{G}$ takes as input only one image instead of a batch. Specifically, let $\hat{\mathbf{x}}_i = \mathcal{G}(\mathbf{x}_i,l_i,l_i^O,\mathbf{x}_i^O)$ be the generated image, starting from $\mathbf{x}_i$ ($\mathbf{x}_i \in \mathbf{D}_{l_i}$) and with desired output domain $l_i^O$ and style image $\mathbf{x}_i^O$.
$\mathcal{G}$ is trained using the combination of three different losses, with the goal of changing the style and the domain of $\mathbf{x}_i$ 
 while preserving its content.

First, we use an {\em adversarial loss} based on the Projection Discriminator (\cite{miyato2018cgans}) ($\mathcal{D_P}$), which is conditioned on labels (domain labels, in our case) and uses a hinge loss:

\vspace{-2mm}
\begin{equation}
    \label{eq.L-cGAN-G}
    \mathcal{L}_{cGAN} (\mathcal{G}) = -\mathcal{D_P}(\hat{\mathbf{x}_i}, l_i^O)
\end{equation}
\vspace{-2mm}
\begin{equation}
    \label{eq.L-cGAN-D}
\begin{split}
    \mathcal{L}_{cGAN} (\mathcal{D_P}) &= \text{max}(0, 1 + \mathcal{D_P}(\hat{\mathbf{x}_i}, l_i^O)) \\
     &+ \text{max}(0, 1 - \mathcal{D_P}(\mathbf{x}_i, l_i))
     \end{split}
\end{equation}

The second loss is the {\em Identity loss} proposed in (\cite{zhu2017unpaired}), which in our framework is implemented as follows:
\vspace{-2mm}
\begin{equation}
\label{eq.ID-loss}
{\cal L}_{ID}(\mathcal{G})=  || \mathcal{G}(\mathbf{x}_i,l_i,l_i,\mathbf{x}_i) - \mathbf{x}_i ||_1.
\end{equation}

In Eq.~\ref{eq.ID-loss}, $\mathcal{G}$ computes an identity transformation, being the input and the output domain and style the same. After that, a pixel-to-pixel $\mathcal{L}_1$ norm is computed.

Finally,  we propose to use a third loss which is based on the rationale that 
the generation process should be {\em equivariant} with respect to a set of simple transformations which preserve the main content of the images (e.g., the foreground object shape). Specifically, we use the set of the affine transformations
$\{ h(\mathbf{x}; \boldsymbol{\theta}) \}$ of image $\mathbf{x}$ which are defined by the parameter $\boldsymbol{\theta}$ ($\boldsymbol{\theta}$ is a 2D transformation matrix). The affine transformation is implemented by a differentiable bilinear kernel as in (\cite{jaderberg2015spatial}). The {\em Equivariance loss} is:

\vspace{-2mm}
\begin{equation}
\label{eq.Geom-loss}
{\cal L}_{Eq}(\mathcal{G})=  || \mathcal{G}(h(\mathbf{x}_i; \boldsymbol{\theta}_i), l_i,l_i^O,\mathbf{x}_i^O)  - 
h(\hat{\mathbf{x}}_i; \boldsymbol{\theta}_i) ||_1.
\end{equation}

In Eq.~\ref{eq.Geom-loss}, for a given image $\mathbf{x}_i$,
we  randomly choose a geometric parameter $\boldsymbol{\theta}_i$ and we apply $h(\cdot; \boldsymbol{\theta}_i)$ to 
$\hat{\mathbf{x}}_i = \mathcal{G}(\mathbf{x}_i,l_i,l_i^O,\mathbf{x}_i^O)$. Then, using the same $\boldsymbol{\theta}_i$,
we apply $h(\cdot; \boldsymbol{\theta}_i)$ to $\mathbf{x}_i$ and we get 
$\mathbf{x}_i' = h(\mathbf{x}_i; \boldsymbol{\theta}_i)$, which is input to $\mathcal{G}$ in order to generate a second  image. The two generated images are finally compared using the $\mathcal{L}_1$ norm. This is a form of self-supervision, in which equivariance to geometric transformations is used to extract semantics. Very recently a similar loss has been proposed in (\cite{hung2019scops}), where equivariance to affine transformations is used for image co-segmentation.

The complete loss for $\mathcal{G}$ is:
\vspace{-2mm}
\begin{equation}
    {\cal L}(\mathcal{G}) = {\cal L}_{cGAN}(\mathcal{G}) + \lambda ({\cal L}_{Eq}(\mathcal{G}) +  {\cal L}_{ID}(\mathcal{G})).
    \label{eqn:gen_equation}
\end{equation}

Note that Eq.~\ref{eq.L-cGAN-G}, \ref{eq.L-cGAN-D} and \ref{eq.Geom-loss} depend on the pair $(\mathbf{x}_i, l_i^O)$: 
This means that 
the supervisory information we effectively use, grows with $O((N+1)^2)$, which is quadratic with respect to a plain strategy in which $N$ different source-to-target generators are trained (Sec.~\ref{Related}).

\textbf{Classifier Training.} 
\label{sec:classifier}
Once $\mathcal{G}$ is trained, we use it to artificially create a labeled training dataset ($T^L$) for the target domain.
 Specifically, for each $S_j$ and each $(\mathbf{x}_i, y_i) \in S_j$, we randomly pick  $\mathbf{x}_t \in T$, which is used as the reference style image, and we generate: $\hat{\mathbf{x}}_i = \mathcal{G}(\mathbf{x}_i,l_i,N+1,\mathbf{x}_t)$, where $N+1$ is fixed and indicates the target domain  ($\mathbf{D}_t$) label (see Sec.~\ref{Overview}).
$(\hat{\mathbf{x}}_i, y_i)$ is added to $T^L$ and the process is iterated.
Note that, in different epochs, for the same $(\mathbf{x}_i, y_i) \in S_j$, we  randomly select a different reference style image $\mathbf{x}_t \in T$. 

Finally, we train a classfier $\mathcal{C}$ on $T^L$
using the cross-entropy loss:
\vspace{-2mm}
\begin{equation}
\label{eq.source-cross-entropy-loss}
{\cal L}_{Cls}(\mathcal{C})=
 - 
 \frac{1}{|T^L|} \sum_{( \hat{\mathbf{x}}_i, y_i) \in T^L}   \log p(y_i | \hat{\mathbf{x}}_i).
\end{equation}

%% file: experiments.tex
\section{Experimental Results}

In this section we describe the experimental setup and then we evaluate our approach using common MSDA datasets. We also present an ablation study in which we separately analyse the impact of each   TriGAN  component. 

\subsection{Datasets} 
\label{sec:datasets}
In our experiments we consider two common domain adaptation benchmarks, namely the Digits-Five benchmark \cite{xu2018deep} and the Office-Caltech dataset~\cite{gong2012geodesic}.

\textbf{Digits-Five}  \cite{xu2018deep} is composed of five  digit-recognition datasets:
USPS \cite{friedman2001elements}, MNIST~\cite{lecun1998gradient}, MNIST-M~\cite{ganin2014unsupervised}, SVHN~\cite{netzer2011reading} and Synthetic numbers datasets~\cite{ganin2016domain} (SYNDIGITS).
SVHN~\cite{netzer2011reading} contains  Google Street View images of real-world house numbers. Synthetic numbers~\cite{ganin2016domain} includes 500K computer-generated digits with different sources of variations (\textit{i.e.} position, orientation, color, blur). USPS \cite{friedman2001elements} is a dataset of digits scanned from U.S. envelopes, MNIST~\cite{lecun1998gradient} is a popular benchmark for digit recognition and MNIST-M~\cite{ganin2014unsupervised} is its colored counterpart. 
We adopt the experimental protocol described in \cite{xu2018deep}: in  each domain the train\slash{}test split is composed of a subset of 25000 images for training and 9000 images for testing. For USPS, the entire dataset is used. 

\textbf{Office-Caltech}~\cite{gong2012geodesic} is a domain-adaptation benchmark, obtained selecting the subset of those $10$  categories which are shared between   Office31 and  Caltech256~\cite{griffin2007caltech}. It contains $2533$ images, about half of which belonging to Caltech256. There are four different domains: Amazon (A), DSLR (D), Webcam (W) and Caltech256 (C).

\begin{table*}[t]
    \centering
	\setlength{\tabcolsep}{2.5pt}
    \begin{tabular}{l|cccccc|c}
        \specialrule{1.5pt}{1pt}{1pt}
        Standards & Models & \makecell{\textbf{mt}, \textbf{up}, \textbf{sv}, \textbf{sy} \\ $\to$ \textbf{mm}} & \makecell{\textbf{mm}, \textbf{up}, \textbf{sv}, \textbf{sy} \\ $\to$ \textbf{mt}} & \makecell{\textbf{mt}, \textbf{mm}, \textbf{sv}, \textbf{sy} \\ $\to$ \textbf{up}} & \makecell{\textbf{mt}, \textbf{up}, \textbf{mm}, \textbf{sy} \\ $\to$ \textbf{sv}} & \makecell{\textbf{mt}, \textbf{up}, \textbf{sv}, \textbf{mm} \\ $\to$ \textbf{sy}} & \textbf{Avg}\\
        \specialrule{1.5pt}{1pt}{1pt}
        \multirow{3}{*}{\makecell{Source \\ Combine}} & Source Only & 63.70$\pm$0.83 & 92.30$\pm$0.91 & 90.71$\pm$0.54 & 71.51$\pm$0.75 & 83.44$\pm$0.79 & 80.33$\pm$0.76 \\
        & DAN\cite{long2015learning} & 67.87$\pm$0.75 & 97.50$\pm$0.62 & 93.49$\pm$0.85 & 67.80$\pm$0.84 & 86.93$\pm$0.93 & 82.72$\pm$0.79 \\
        & DANN\cite{ganin2014unsupervised} & 70.81$\pm$0.94 & 97.90$\pm$0.83 & 93.47$\pm$0.79 & 68.50$\pm$0.85 & 87.37$\pm$0.68 & 83.61$\pm$0.82 \\
        \hline
        \multirow{8}{*}{\makecell{Multi-\\Source}}& Source Only & 63.37$\pm$0.74 & 90.50$\pm$0.83 & 88.71$\pm$0.89 & 63.54$\pm$0.93 & 82.44$\pm$0.65 & 77.71$\pm$0.81 \\
        & DAN\cite{long2015learning} & 63.78$\pm$0.71 & 96.31$\pm$0.54 & 94.24$\pm$0.87 & 62.45$\pm$0.72 & 85.43$\pm$0.77 & 80.44$\pm$0.72 \\
        & CORAL\cite{sun2016return} & 62.53$\pm$0.69 & 97.21$\pm$0.83 & 93.45$\pm$0.82 & 64.40$\pm$0.72 & 82.77$\pm$0.69 & 80.07$\pm$0.75 \\
        & DANN\cite{ganin2014unsupervised} & 71.30$\pm$0.56 & 97.60$\pm$0.75 & 92.33$\pm$0.85 & 63.48$\pm$0.79 & 85.34$\pm$0.84 & 82.01$\pm$0.76 \\
        & ADDA\cite{tzeng2017adversarial} & 71.57$\pm$0.52 & 97.89$\pm$0.84 & 92.83$\pm$0.74 & 75.48$\pm$0.48 & 86.45$\pm$0.62 & 84.84$\pm$0.64 \\
        & DCTN\cite{xu2018deep} & 70.53$\pm$1.24 & 96.23$\pm$0.82 & 92.81$\pm$0.27 & 77.61$\pm$0.41 & 86.77$\pm$0.78 & 84.79$\pm$0.72 \\
        & $\textrm{M}^{3}\textrm{SDA}$\cite{peng2018moment} & \underline{72.82}$\pm$1.13 & \textbf{98.43}$\pm$0.68 & \textbf{96.14}$\pm$0.81 & \underline{81.32}$\pm$0.86 & \underline{89.58}$\pm$0.56 & \underline{87.65}$\pm$0.75 \\
        & StarGAN~\cite{choi2018stargan} & 44.71$\pm$1.39 & \underline{96.26}$\pm$0.62 & 55.32$\pm$3.71 & 58.93$\pm$1.95 & 63.36$\pm$2.41 & 63.71$\pm$2.01 \\
        & TriGAN (Ours) & \textbf{83.20}$\pm$0.78 & \underline{97.20}$\pm$0.45 & \underline{94.08}$\pm$0.92 & \textbf{85.66}$\pm$0.79 & \textbf{90.30}$\pm$0.57 & \textbf{90.08}$\pm$0.70 \\
        \specialrule{1.5pt}{1pt}{1pt}
    \end{tabular}
    \caption{Classification accuracy (\%) on \textbf{Digits-Five}. \textit{MNIST-M, MNIST, USPS, SVHN, }\textit{Synthetic Digits} are abbreviated as \textbf{mm}, \textbf{mt}, \textbf{up}, \textbf{sv} and \textbf{sy} respectively. Best number is in bold and second best is underlined.}
    \label{tab:digits_sota}
\end{table*}

\subsection{Experimental Setup}

 For lack of space, we provide the architectural details of our generator $\mathcal{G}$ and discriminator $\mathcal{D}_{\mathcal{P}}$ networks  in the Supplementary Material. We train TriGAN for 100 epochs using the Adam optimizer \cite{kingma2014adam} with the learning rate set to 1e-4 for  $\mathcal{G}$ and 4e-4 for  $\mathcal{D}_{\mathcal{P}}$ as in \cite{heusel2017gans}. 
 The loss weighing factor $\lambda$ in Eqn. \ref{eqn:gen_equation} is set to 10 as in \cite{zhu2017unpaired}. 

In the Digits-Five experiments we use a mini-batch of size 256 for TriGAN training. Due to the difference in image resolution and  image channels, the images of all the domains are converted to 32 $\times$ 32 RGB. For a fair comparison, for the final target classifier $\mathcal{C}$ we use exactly the same network architecture used in \cite{ganin2016domain,peng2018moment}.

In the Office-Caltech10 experiments we downsample the images to 164 $\times$ 164  to accommodate more samples in a mini-batch. We use a mini-batch of size 24 for training with 1 GPU. For the back-bone target classifier $\mathcal{C}$ we use the ResNet101 \cite{he2016deep} architecture  used by \cite{peng2018moment}. The weights  are initialized with a network pre-trained  on the ILSVRC-2012 dataset \cite{russakovsky2015imagenet}. In our experiments we remove the output layer and we replace it with a randomly initialized fully-connected layer with 10 logits, one for each class of the Office-Caltech10 dataset. $\mathcal{C}$ is trained with Adam with an initial learning rate of 1e-5 for the randomly initialized last layer and 1e-6 for all other layers. 
In this setting we also include $\{S_{j} \}_{j=1}^N$ in $T^L$ for training the classifier $\mathcal{C}$. 


\subsection{Results}
In this section we quantitatively analyse TriGAN. In the Supplementary Material we show some qualitative results for \textbf{Digits-Five} and \textbf{Office-Caltech10}.

\subsubsection{Comparison with State-of-the-Art Methods}
Tab.~\ref{tab:digits_sota} and Tab.~\ref{tab:office_caltech_sota} show the results on the Digits-Five and the Office-Caltech10 datset, respectively. 
Table \ref{tab:digits_sota} shows that TriGAN achieves an average accuracy of {90.08}\% which is higher than all other methods. $\textrm{M}^{3}\textrm{SDA}$ is better in the \textbf{mm}, \textbf{up},  \textbf{sv}, \textbf{sy} $\to$ \textbf{mt} and in the \textbf{mt}, \textbf{mm}, \textbf{sv}, \textbf{sy} $\to$ \textbf{up} settings, where  TriGAN is the second best. In all the other settings, TriGAN outperforms all the other approaches. As an example, in the \textbf{mt}, \textbf{up}, \textbf{sv}, \textbf{sy} $\to$ \textbf{mm} setting, TriGAN is better than the second best method $\textrm{M}^{3}\textrm{SDA}$ by a significant margin of 10.38\%. 
In the same table we also show the results obtained when we replace TriGAN  with  
StarGAN~\cite{choi2018stargan}, which is another ``universal'' image translator. Specifically, we use  StarGAN to
 generate synthetic target images and then we train the target classifier using  the same  protocol described in Sec.~\ref{sec:net_train}. 
 The corresponding results in Table \ref{tab:digits_sota} show that
 StarGAN, despite to be known to work well for aligned face translation,  drastically fails when used in this UDA scenario.

Finally, we also  use  Office-Caltech10, which is considered to be difficult for reconstruction-based GAN methods because of the high-resolution  images. Although the dataset is quite saturated, TriGAN achieves a classification accuracy of {97.0}\%, outperforming all the other methods and beating the previous state-of-the-art approach ($\textrm{M}^{3}\textrm{SDA}$) by a margin of 0.6\% on average (see Tab.~\ref{tab:office_caltech_sota}). 

\begin{table}[!h]
    \centering
	\setlength{\tabcolsep}{3.5pt}
    \begin{tabular}{l|ccccc|c}
        \specialrule{1.5pt}{1pt}{1pt}
        \thead{Standards} & \thead{Models} & \thead{$\to$ W} & \thead{$\to$ D} & \thead{$\to$ C} & \thead{$\to$ A} & \thead{Avg}\\
        \hline
        \multirow{2}{*}{\makecell{Source\\Combine}} & Source only & 99.0 & 98.3 & 87.8 & 86.1 & 92.8\\
        & DAN \cite{long2015learning}& 99.3 & 98.2 & 89.7 & 94.8 & 95.5\\
        \hline
        \multirow{5}{*}{\makecell{Multi-\\Source}} & Source only & 99.1 & 98.2 & 85.4 & 88.7 & 92.9\\
        & DAN \cite{long2015learning} & 99.5 & 99.1 &  89.2 &  91.6 & 94.8 \\
        & DCTN \cite{xu2018deep} & 99.4 & 99.0 & 90.2 & 92.7 & 95.3\\
        & $\textrm{M}^{3}\textrm{SDA}$ \cite{peng2018moment} & \underline{99.5} & \underline{99.2} & \underline{92.2} & \underline{94.5} & \underline{96.4}\\
        & StarGAN \cite{choi2018stargan} & 99.6 & \textbf{100.0} & 89.3 & 93.3 & 95.5 \\
        & TriGAN (Ours) & \textbf{99.7} & \textbf{100.0} & \textbf{93.0} & \textbf{95.2} & \textbf{97.0} \\
        \specialrule{1.5pt}{1pt}{1pt}
    \end{tabular}
    \caption{Classification accuracy (\%) on \textbf{Office-Caltech10}.}
    \label{tab:office_caltech_sota}
\end{table}

\subsubsection{Ablation Study}

In this section we  analyse the different components of our  method and study in isolation their impact on the final accuracy.
Specifically, we use the Digits-Five dataset and the following models: i) Model \textbf{A}, which is  our full model containing the following components: \textit{IWT}, \textit{DWT}, \textit{cDWT}, \textit{AdaIWT} and $\mathcal{L}_{Eq}$. ii) Model \textbf{B}, which is similar to Model \textbf{A} except we replace  $\mathcal{L}_{Eq}$ with the cycle-consistency loss $\mathcal{L}_{Cycle}$ of CycleGAN \cite{zhu2017unpaired}.  iii) Model \textbf{C}, where we replace \textit{IWT}, \textit{DWT}, \textit{cDWT} and \textit{AdaIWT} of Model \textbf{A} with \textit{IN} \cite{DBLP:journals/corr/UlyanovVL16}, \textit{BN} \cite{ioffe2015batch}, conditional Batch Normalization (\textit{cBN}) \cite{DBLP:journals/corr/DumoulinSK16} and Adaptive Instance Normalization (\textit{AdaIN}) \cite{huang2018multimodal}. This comparison highlights the difference between 
 feature whitening and feature standardisation. iv) Model \textbf{D}, which ignores the style factor. Specifically, in Model \textbf{D}, the blocks related to the style factor, i.e., the \textit{IWT} and the \textit{AdaIWT} blocks, are replaced by \textit{DWT} and \textit{cDWT} blocks, respectively. v) Model \textbf{E}, in which the style path differs from Model \textbf{A} in the way the style is applied to the domain-specific representation. Specifically, we remove the MLP $\mathcal{F}(.)$ and
we directly apply ($\boldsymbol{\mu}_{X_i^O}, \boldsymbol{W}_{X_i^O}^{-1}$). vi) Finally, Model \textbf{F} represents no-domain assumption (e.g. the DWT and cDWT blocks are replaced with standard WC blocks).
    
\begin{table}[h!]
    \centering
	\setlength{\tabcolsep}{2pt}
    \begin{tabular}{lc|c}
        \specialrule{1.5pt}{1pt}{1pt}
         Model & Description & \makecell{Avg. Accuracy (\%) \\ (Difference)} \\
         \hline
         \textbf{A} & TriGAN (full method) & \textbf{90.08}\\
         \textbf{B} & \makecell{Replace Equivariance  Loss \\ with Cycle Loss} & 88.38 (\textcolor{red}{-1.70})\\
         \textbf{C} & \makecell{Replace Whitening with \\ Feature Standardisation} & 89.39 (\textcolor{red}{-0.68})\\
         \textbf{D} & No Style Assumption & 88.32 (\textcolor{red}{-1.76}) \\
         \textbf{E} & \makecell{Applying style directly \\ instead of style path} & 88.36 (\textcolor{red}{-1.71}) \\
         \textbf{F} & No Domain Assumption & 89.10 (\textcolor{red}{-0.98})\\
         \specialrule{1.5pt}{1pt}{1pt}
    \end{tabular}
    \caption{An analysis of the main TriGAN components using Digits-Five.}
    \label{tab:ablation}
\end{table}

Tab. \ref{tab:ablation} shows that Model \textbf{A} outperforms all the ablated models. Model \textbf{B} shows that $\mathcal{L}_{Cycle}$ is detrimental for the accuracy because  $\mathcal{G}$ may focus on meaningless information to reconstruct back the image. Conversely, the affine transformations used in case of $\mathcal{L}_{Eq}$, force  $\mathcal{G}$ to  focus on the shape (i.e., the content) of the images. Also Model \textbf{C} is outperformed by model \textbf{A}, demonstrating the importance of feature whitening over feature standardisation, corroborating the findings of \cite{roy2019unsupervised} in a pure-discriminative setting. Moreover, the no-style assumption in Model \textbf{D} hurts the classification accuracy by a margin of 1.76\% when compared with  Model \textbf{A}. We believe this is due to the fact that, when only domain-specific latent factors are modeled but instance-specific style information is missing in the image translation process, then the diversity of the translations decreases, consequently reducing the final accuracy
(see the role of the randomly picked $\mathbf{x}_t \in T$, in Sec.~\ref{sec:classifier}). Model \textbf{E} shows the need of using the proposed style path. Finally,  Model \textbf{F}  shows that having a separate factor for domain yields a better performance.
Note that the ablation analysis in Tab. \ref{tab:ablation} is done 
by  removing a single component from the full model \textbf{A}, and the marginal difference with Model \textbf{A} shows that all the components are important. On the other hand, simultaneously removing  all the components makes our  model become similar to StarGAN, where there is no style information and where the domain information is not ``whitened'' but provided as input to the network. As shown in Table \ref{tab:digits_sota},  our full model drastically outperfoms a StarGAN-based generative MSDA approach.

\subsubsection{Multi domain image-to-image translation}
\begin{figure}[!h]
    \centering
    \includegraphics[width=\columnwidth]{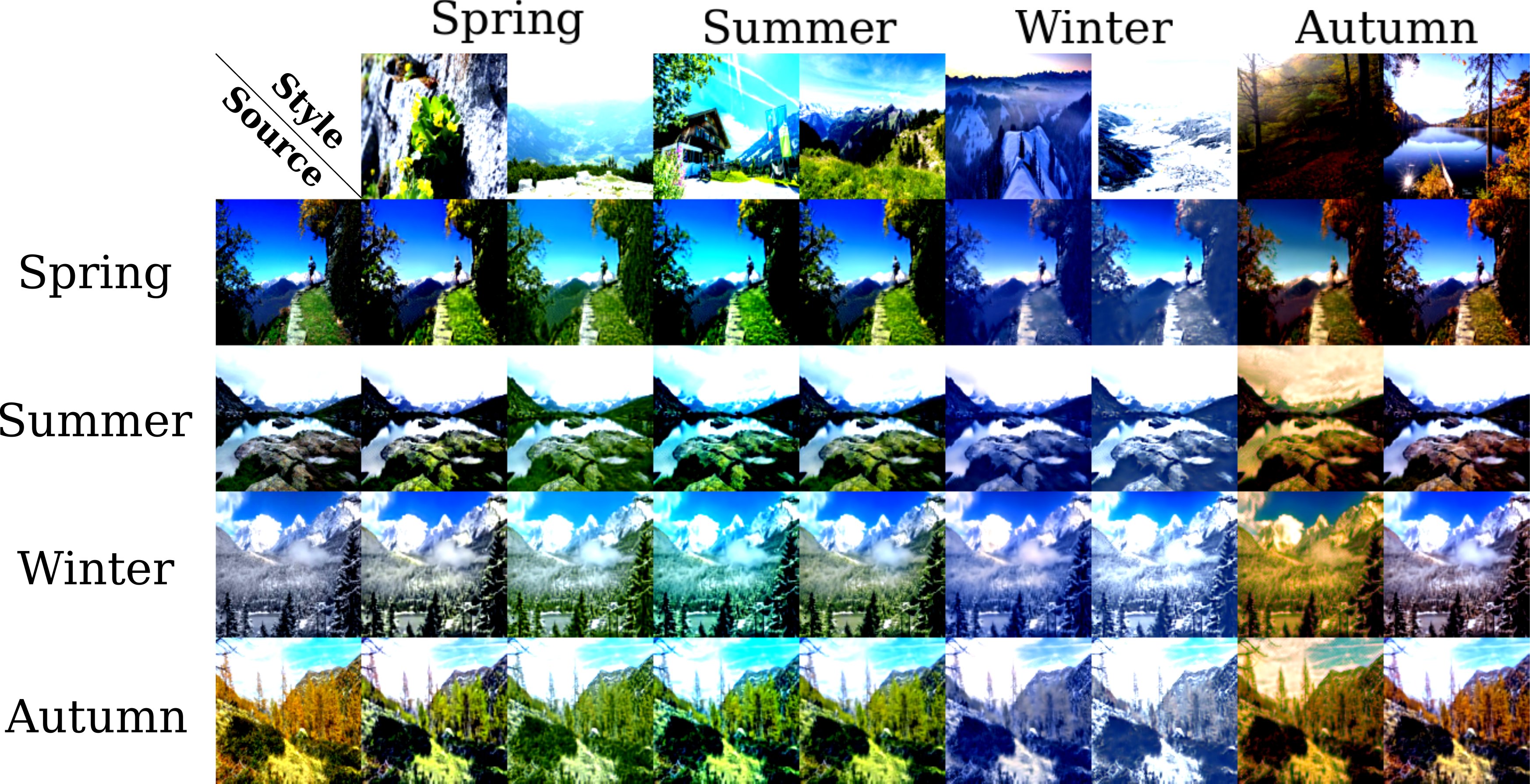}
  \caption{Some example images generated by TriGAN across different domains (i.e., seasons). We show two generated images for each  domain combination. This figure is reported also in the Supplementary
Material with a higher resolution.}
  \label{fig:sample_alps_generation}
  \vspace{-0.5cm}
\end{figure}

Our proposed generator can be used for a pure generative (non-UDA), multi-domain image-to-image translation task. We conduct experiments on the Alps Seasons dataset \cite{anoosheh2018combogan} which consists of images of Alps mountains with 4 different domains (corresponding to 4 seasons).
Fig.~\ref{fig:sample_alps_generation} shows some images generated using our generator. For this experiment we compare our generator with StarGAN~\cite{choi2018stargan} using the FID~\cite{heusel2017gans} metrics. FID measures the realism of the generated images (the lower the better). The FID scores are computed considering all the real samples in the target domain and generating an equivalent number of synthetic images in the target domain. Tab.~\ref{tab:fid_alps} shows that the TriGAN FID scores are significantly lower than the StarGAN scores. This further highlights 
that decoupling the style and the domain and using $WC$-based layers to progressively ``whiten'' and ``color'' the image statistics, yields to a more realistic cross-domain image translation than using domain labels as input as in the case of StarGAN.

\begin{table}[h!]
    \centering
	\setlength{\tabcolsep}{2pt}
    \begin{tabular}{lcccc}
    \specialrule{1.5pt}{1pt}{1pt}
         Target & $\rightarrow$Winter & $\rightarrow$Summer & $\rightarrow$Spring & $\rightarrow$Autumn  \\
         \hline
         StarGAN~\cite{choi2018stargan} & 148.45 & 180.36 & 175.40 &145.24 \\
         TriGAN (Ours) & 41.03 & 38.59 & 40.75 & 32.71\\
         \specialrule{1.5pt}{1pt}{1pt}
    \end{tabular}
    \caption{Alps Seasons, FID scores: Comparing TriGAN with StarGAN~\cite{choi2018stargan}.}
    \label{tab:fid_alps}
    \vspace{-0.5cm}
\end{table}

%% file: arxiv-supp.tex
\renewcommand{\thesection}{\Alph{section}}
\setcounter{section}{0}

\section{Additional Multi-Source Results}
\label{sec:addln_exps}
\begin{figure*}[h!]
  \centering\includegraphics[width=0.95\linewidth]{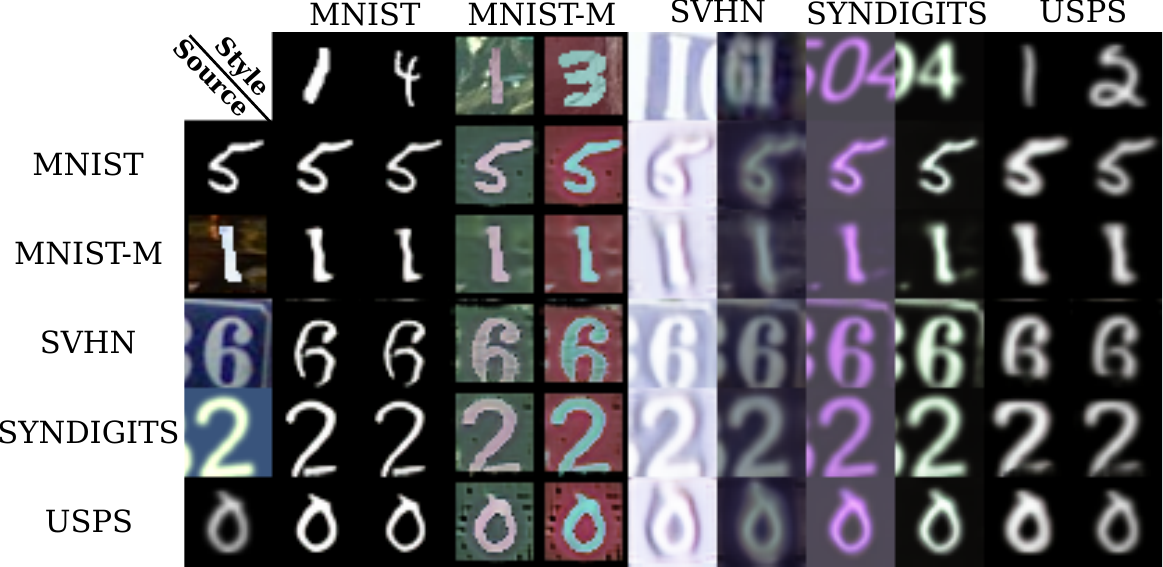}
  \caption{Generations of our $\mathcal{G}$ across different domains of Digits-Five. Leftmost column shows the \textit{source} images, one from each domain and the topmost row shows the \textit{style} image from the target domain, two from each domain.}
  \label{fig:sample_generation_digits}
\end{figure*}

\begin{figure*}[h!]
  \centering\includegraphics[width=0.95\linewidth]{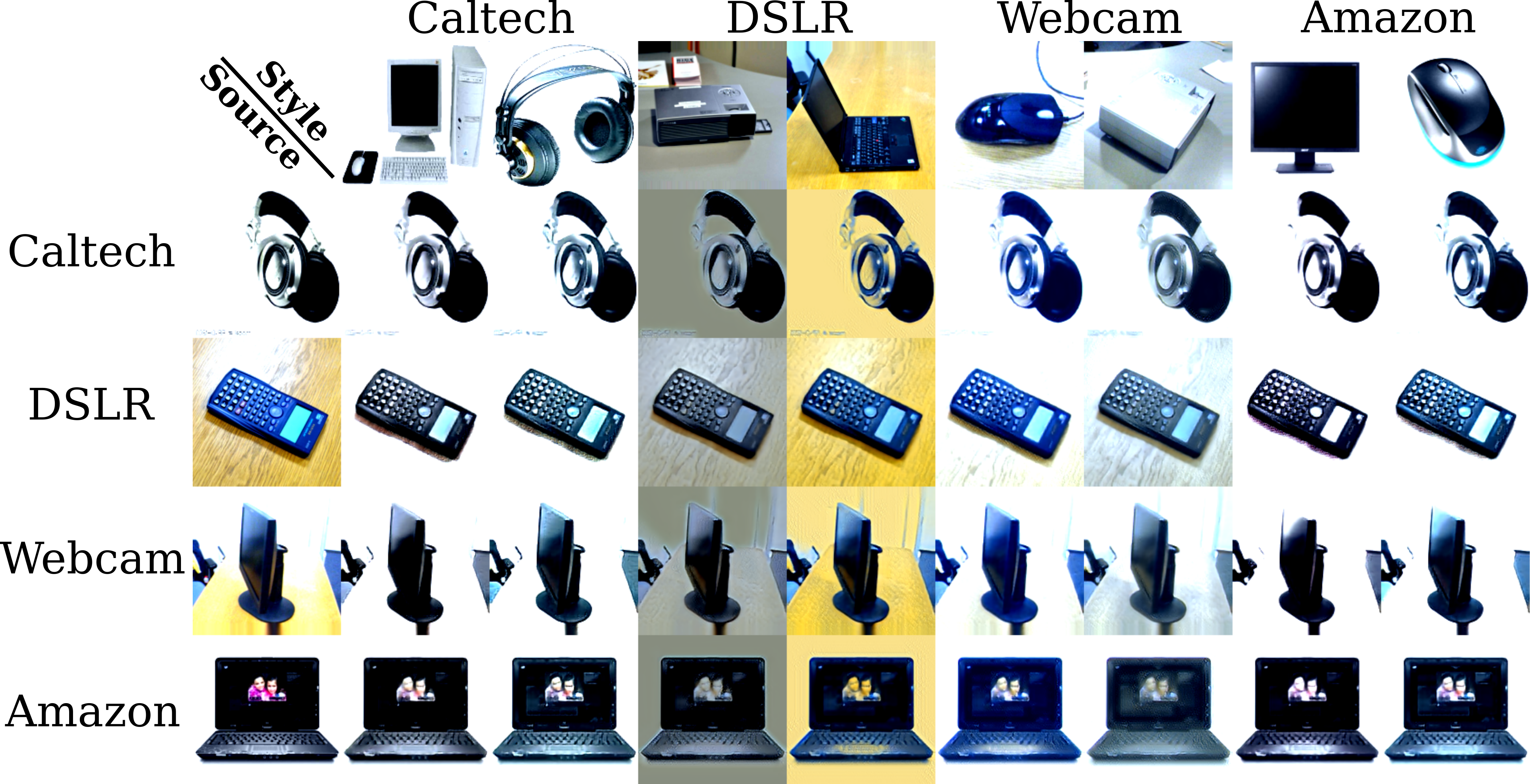}
  \caption{Generations of our $\mathcal{G}$ across different domains of Office-Caltech10. Leftmost column shows the \textit{source} images, one from each domain and the topmost row shows the \textit{style} image from the target domain, two from each domain.}
  \label{fig:sample_generation_office}
\end{figure*}

\begin{figure*}[h!]
  \centering\includegraphics[width=0.95\linewidth]{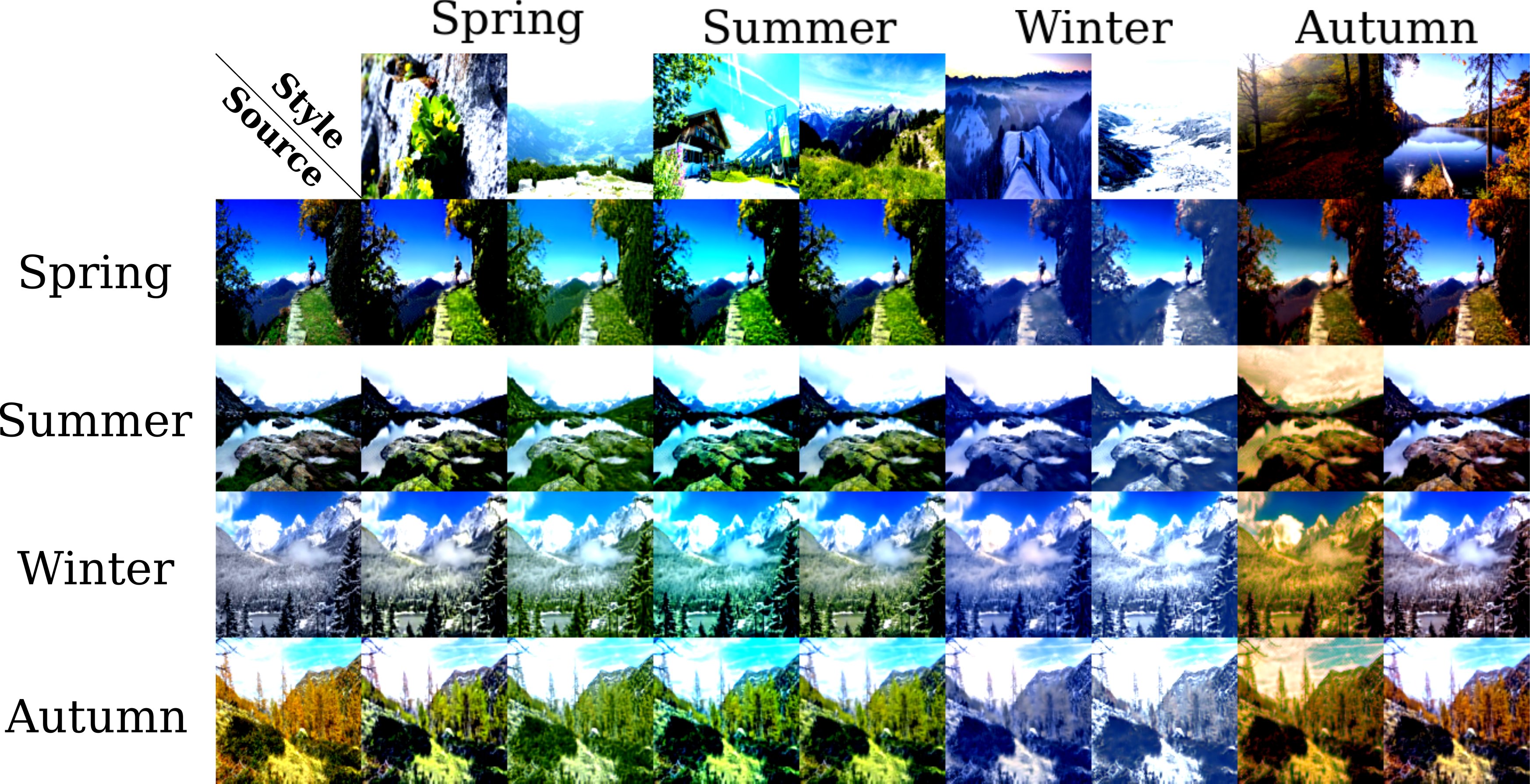}
  \caption{Generations of our $\mathcal{G}$ across different domains of Alps dataset. Leftmost column shows the \textit{source} images, one from each domain and the topmost row shows the \textit{style} image from the target domain, two from each domain.}
  \label{fig:sample_generation_alps}
\end{figure*}

Some sample translations of our $\mathcal{G}$ are shown in Figs.~\ref{fig:sample_generation_digits},~\ref{fig:sample_generation_office},~\ref{fig:sample_generation_alps}. For example, in Fig.~\ref{fig:sample_generation_digits} when the SVHN digit \enquote{six} with side-digits is translated to MNIST-M the \textit{cDWT} blocks re-projects it to MNIST-M domain (i.e., single digit without side-digits) and the \textit{AdaIWT} block applies the instance-specific style of the digit \enquote{three} (i.e., blue digit with red background) to yield a blue \enquote{six} with red background. Similar trends are also observed in Fig. \ref{fig:sample_generation_office}. 

\section{Implementation details}
\label{sec:impl_det}
In this section we provide the architecture details of the TriGAN generator $\mathcal{G}$ and the discriminator $\mathcal{D}_{\mathcal{P}}$.\\

\begin{figure}[h]
    \centering
    \begin{subfigure}{0.3\columnwidth}
    \centering
          \includegraphics[width=0.7\columnwidth]{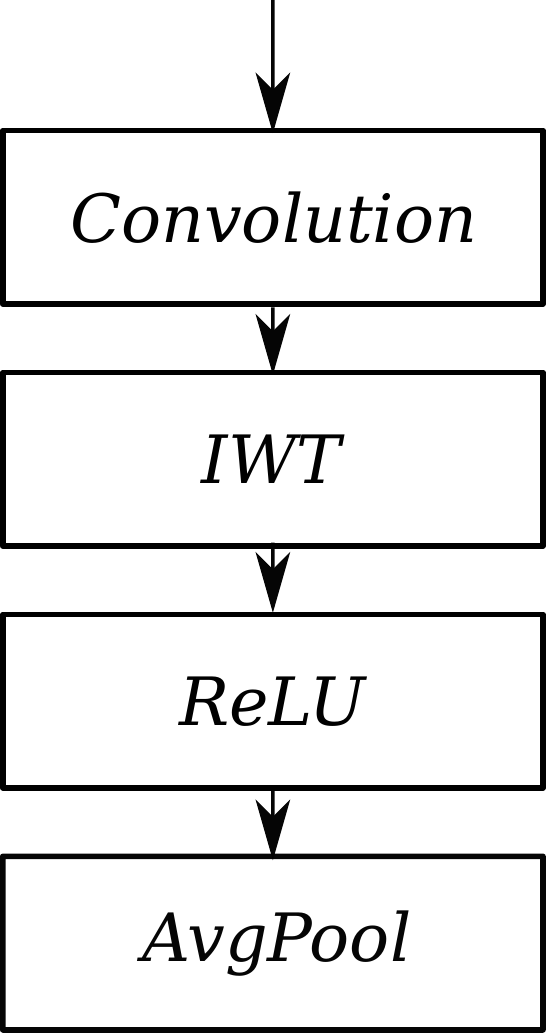}
          \label{fig.iwt}
          \caption{\textbf{IWT} block}
    \end{subfigure}
    \begin{subfigure}{0.3\columnwidth}
    \centering
          \includegraphics[width=0.85\columnwidth]{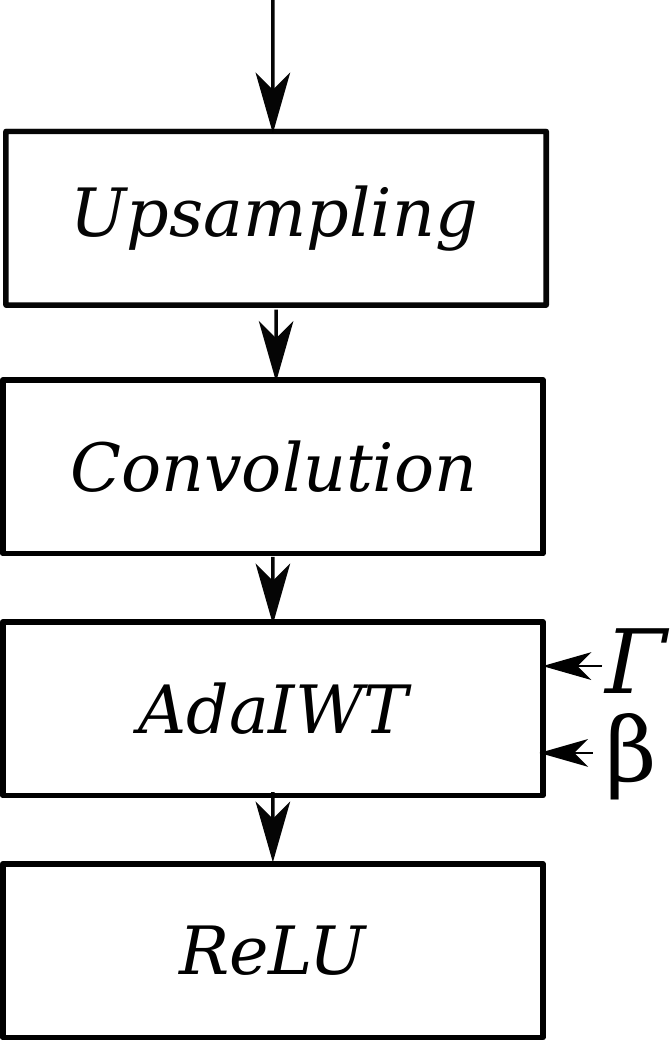}
          \label{fig.dwt}
          \caption{\textbf{AdaIWT} block}  
    \end{subfigure}
    \begin{subfigure}{0.35\columnwidth}	
    \centering
          \includegraphics[width=1.1\columnwidth]{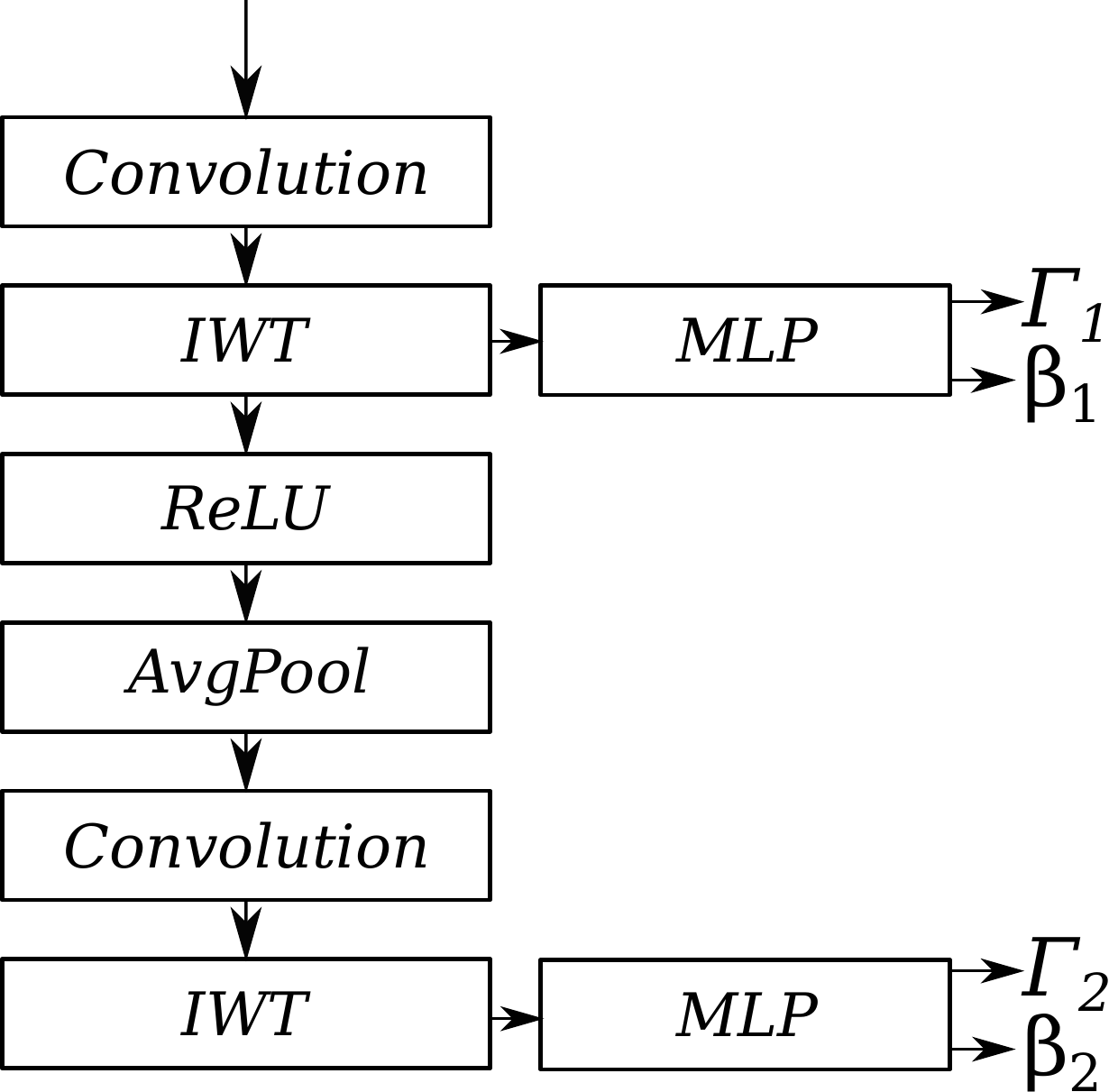}
          \label{fig.style}
          \caption{\textbf{Style Path}}  
    \end{subfigure}
  \caption{A schematic representation of the (a) \textbf{IWT} block; (b) \textbf{AdaIWT} block; and (c) \textbf{Style Path}.}
  \vspace{-0.5cm}
  \label{fig:iwt_blocks}
\end{figure}

\noindent \textbf{Instance Whitening Transform} (\textbf{IWT}) \textbf{blocks}. As shown in Fig~\ref{fig:iwt_blocks} (a) each \textbf{IWT} block is a sequence composed of:  $Convolution_{k \times k} - IWT - ReLU - AvgPool_{m \times m}$, where $k$ and $m$ denote the kernel sizes. There are two \textbf{IWT} blocks in the $\mathcal{E}$. 
In the first \textbf{IWT} block we use $k = 5$ and $m= 2$, and in the second we use $k = 3$ and $m= 2$.

\noindent \textbf{Adaptive Instance Whitening} (\textbf{AdaIWT}) \textbf{blocks}. The \textbf{AdaIWT} blocks are analogous to the \textbf{IWT} blocks except from the \textit{IWT}  which is replaced by the \textit{AdaIWT}. The \textbf{AdaIWT} block is a sequence:  $Upsampling_{m \times m} - Convolution_{k \times k} - AdaIWT - ReLU$, where $m = 2$ and $k = 3$.  \textit{AdaIWT}  also takes as  input the coloring parameters ($\boldsymbol{\Gamma}$, $\boldsymbol{\beta}$) 
(See Sec.~3.2.3) and
 Fig.~\ref{fig:iwt_blocks} (b)). Two \textbf{AdaIWT} blocks are consecutively used  in  $\mathcal{D}$. The last \textbf{AdaIWT} block is followed by a $Convolution_{5 \times 5}$ layer.\\

\noindent \textbf{Style Path}. The \textbf{Style Path}  is composed of: $Convolution_{5 \times 5} - (IWT - MLP) - ReLU - AvgPool_{2 \times 2} - Convolution_{3 \times 3} - (IWT - MLP)$ (Fig.~\ref{fig:iwt_blocks} (c)). The output of the \textbf{Style Path} is $(\boldsymbol{\beta}_{1} \Vert \boldsymbol{\Gamma}_{1})$ and  $(\boldsymbol{\beta}_{2} \Vert \boldsymbol{\Gamma}_{2})$, which are input to the  second and the first \textbf{AdaIWT} blocks, respectively (see Fig.~\ref{fig:iwt_blocks} (b)). The $MLP$ is composed of five fully-connected layers with 256, 128, 128, 256 neurons, 
with the last fully-connected layer having a number of neurons equal to the cardinality of the coloring parameters $(\boldsymbol{\beta} \Vert \boldsymbol{\Gamma})$.

\begin{figure}[h]
    \centering
    \begin{subfigure}{0.4\columnwidth}
    \centering
          \includegraphics[width=0.5\columnwidth]{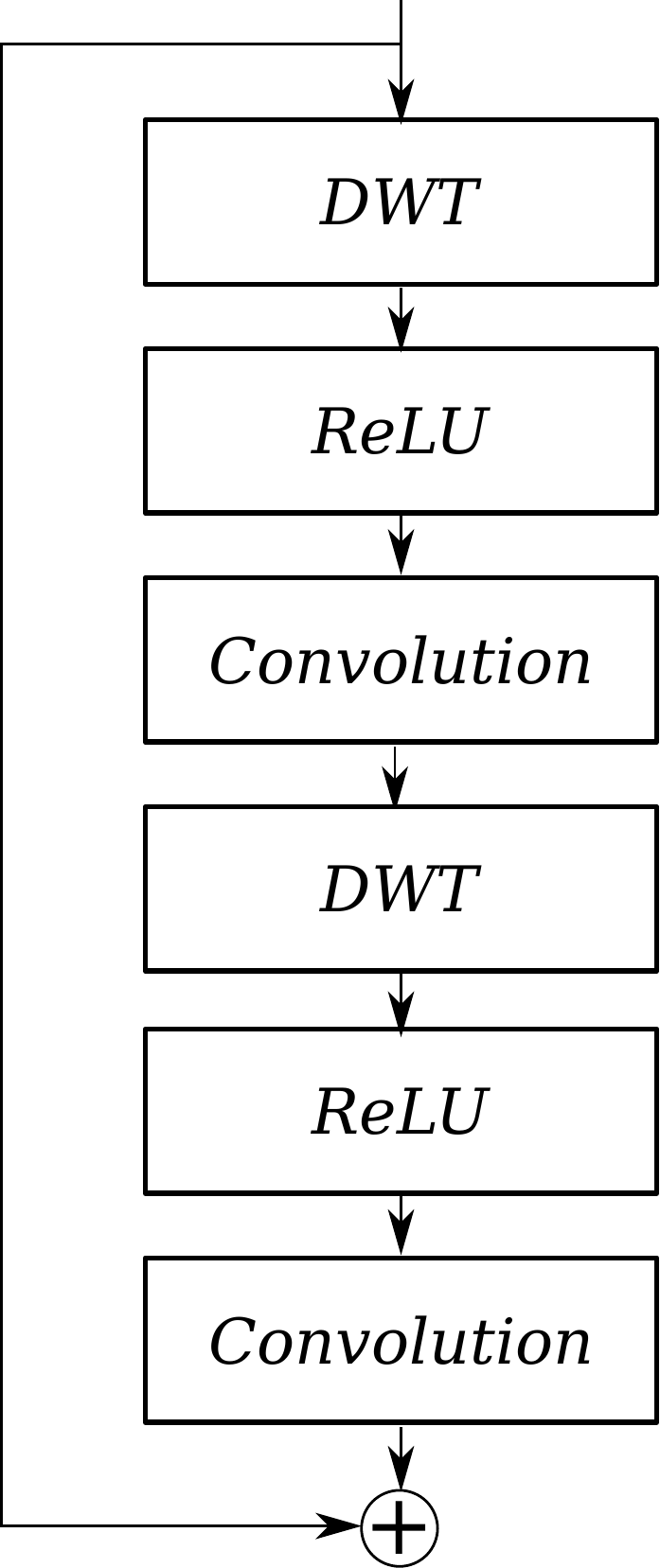}
          \label{fig.cdwt}
          \caption{\textbf{DWT} block}
    \end{subfigure}
    \begin{subfigure}{0.4\columnwidth}	
    \centering
          \includegraphics[width=0.5\columnwidth]{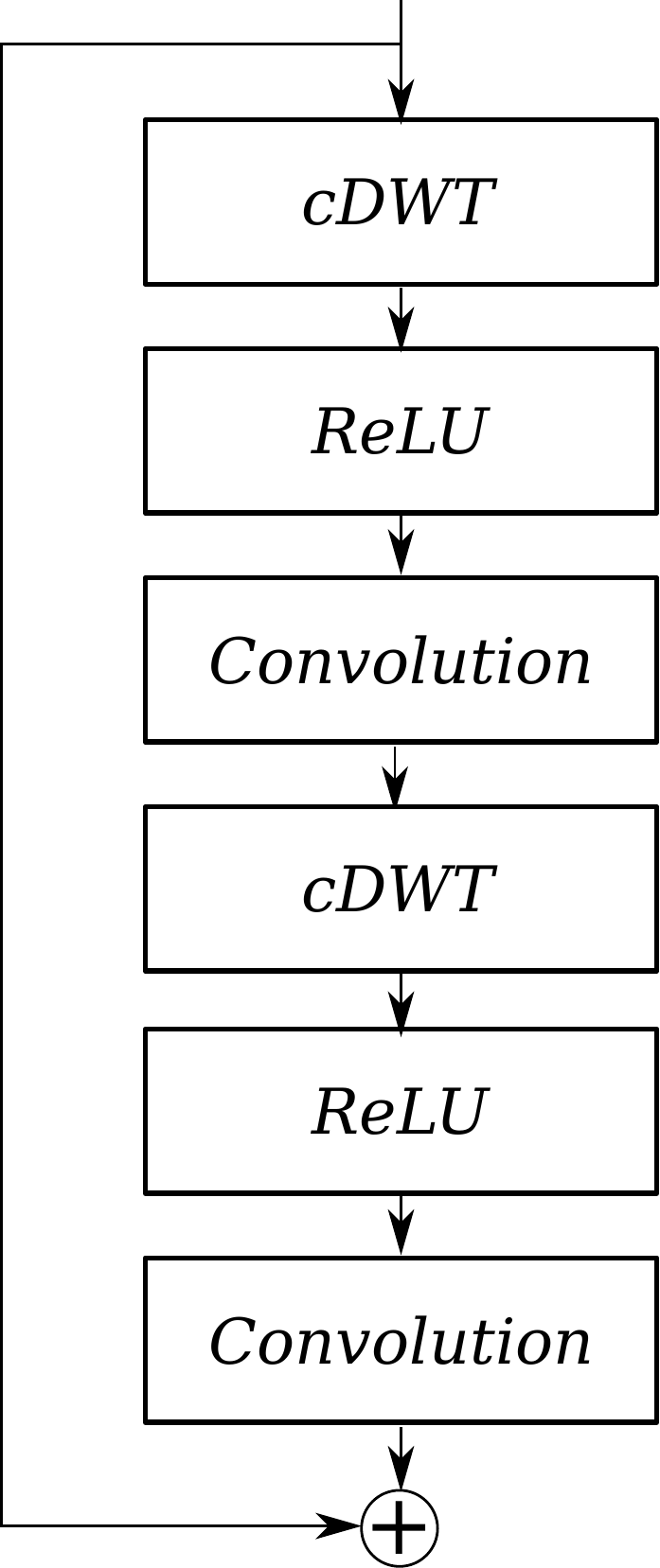}
          \label{fig.adaIWT}
          \caption{\textbf{cDWT} block}  
    \end{subfigure}
  \caption{Schematic representation of (a) \textbf{DWT} block; and (b) \textbf{cDWT} block.}
  \label{fig:dwt_blocks}
\end{figure}

\noindent \textbf{Domain Whitening Transform} (\textbf{DWT}) \textbf{blocks}. The schematic representation of a \textbf{DWT} block is shown in Fig.~\ref{fig:dwt_blocks} (a). For the \textbf{DWT} blocks we adopt a residual-like structure  \cite{he2016deep}: $DWT - ReLU - Convolution_{3 \times 3} - DWT - ReLU - Convolution_{3 \times 3}$. We also add identity shortcuts in the \textbf{DWT} residual blocks to aid the training process.

\noindent \textbf{Conditional Domain Whitening Transform} (\textbf{cDWT}) \textbf{blocks}. The proposed \textbf{cDWT} blocks are schematically shown in Fig.~\ref{fig:dwt_blocks} (b). Similarly to a \textbf{DWT} block, a \textbf{cDWT} block contains the following layers: $cDWT - ReLU - Convolution_{3 \times 3} - cDWT - ReLU - Convolution_{3 \times 3}$. Identity shortcuts are also used in the \textbf{cDWT} residual blocks.

\begin{figure}[h]
    \centering
    \includegraphics[width=0.7\columnwidth]{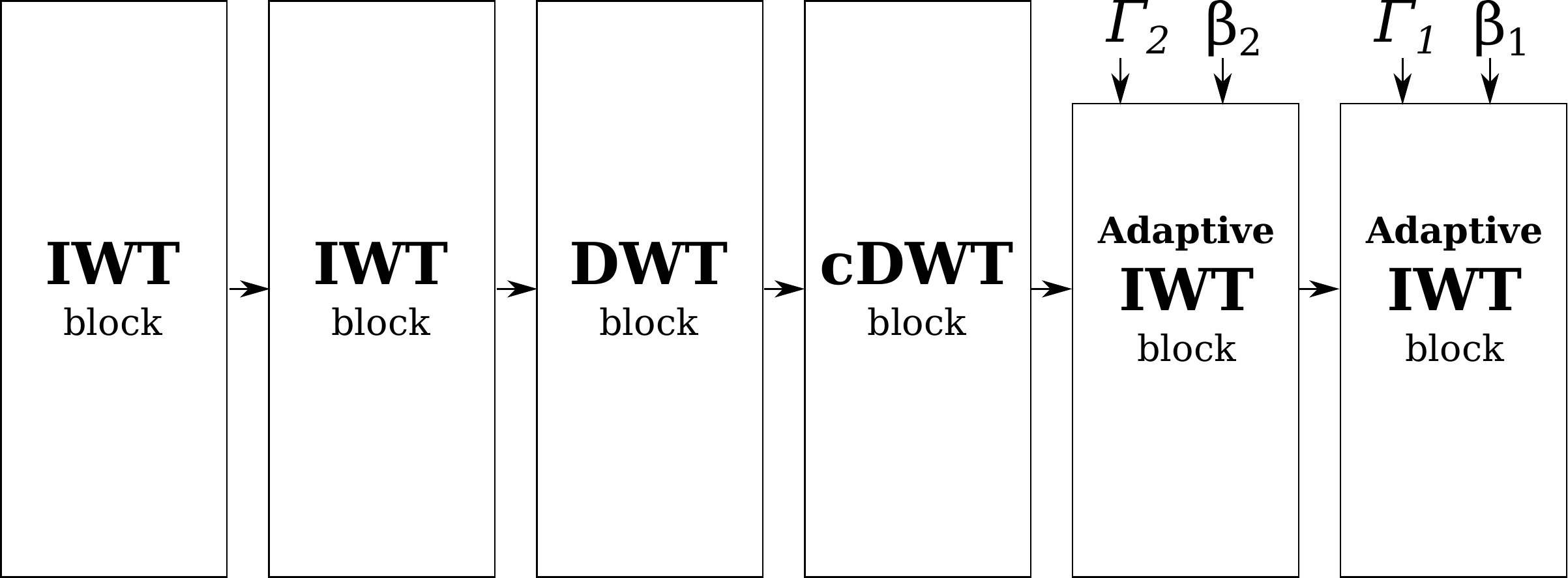}
  \caption{Schematic representation of the Generator $\mathcal{G}$ block.}
  \label{fig:generator_schematic}
\end{figure}

All the above blocks are assembled to construct $\mathcal{G}$, as shown in Fig.~\ref{fig:generator_schematic}. Specifically,  $\mathcal{G}$ contains two \textbf{IWT} blocks, one \textbf{DWT} block, one \textbf{cDWT} block and two \textbf{AdaIWT} blocks. It also contains the \textbf{Style Path} and 2 \textit{$Convolution_{5 \times 5}$} (one before the first \textbf{IWT} block and another after the last \textbf{AdaIWT} block), which  is omitted in Fig.~\ref{fig:generator_schematic} for the sake of clarity.    \{$\boldsymbol{\Gamma}_{1}, \boldsymbol{\beta}_{1}, \boldsymbol{\Gamma}_{2}, \boldsymbol{\beta}_{2}$\} are computed using the \textbf{Style Path}.

\begin{figure}[h]
    \centering
    \includegraphics[width=0.4\columnwidth]{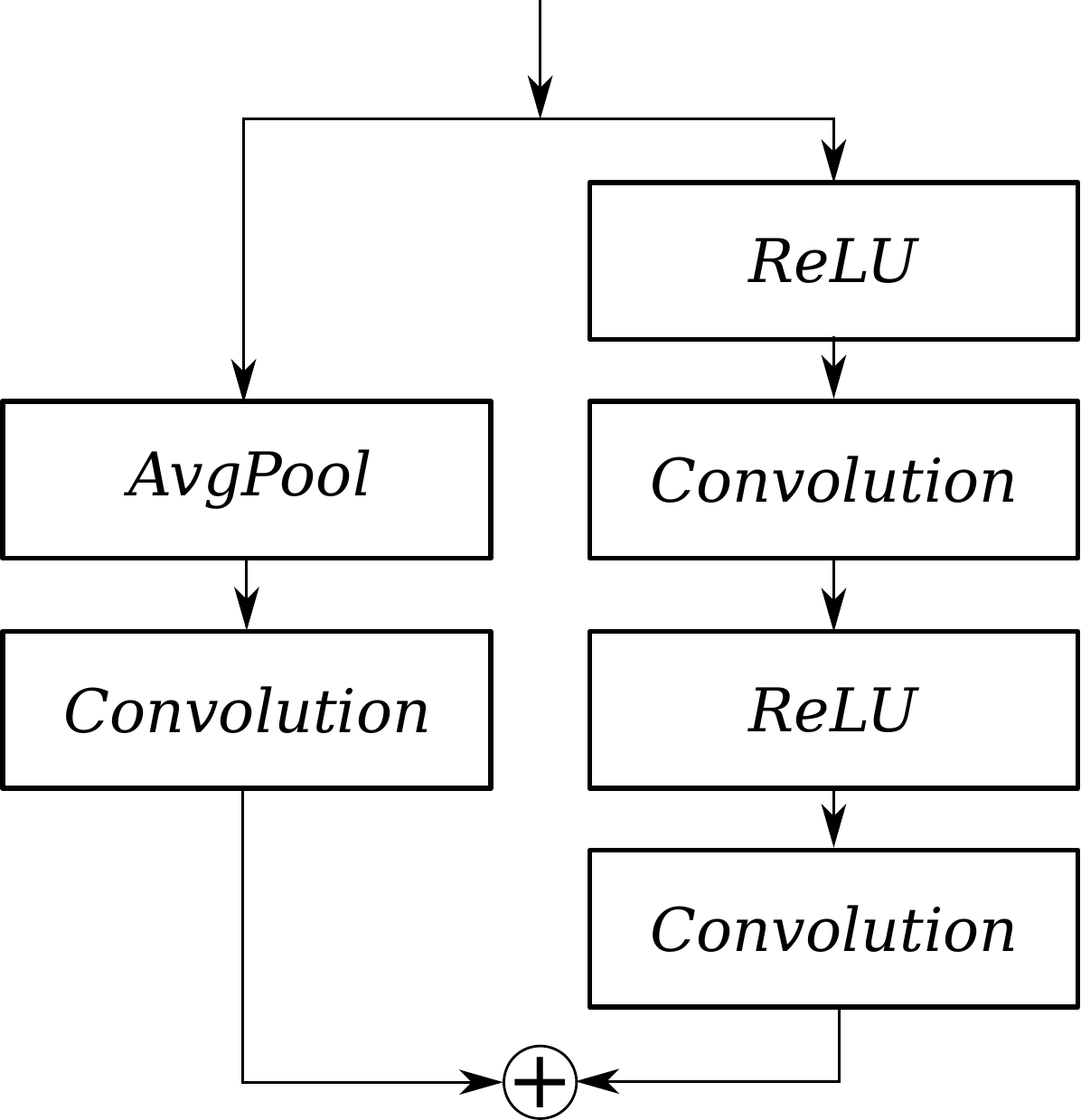}
  \caption{Schematic representation of the Discriminator $\mathcal{D_P}$ block.}
  \label{fig:discriminator_schematic}
\end{figure}

For the discriminator $\mathcal{D}_{\mathcal{P}}$ architecture we use a Projection Discriminator~\cite{miyato2018cgans}. In $\mathcal{D}_{\mathcal{P}}$ we use projection shortcuts instead of identity shortcuts. In Fig~\ref{fig:discriminator_schematic} we  schematically show a discriminator block. $\mathcal{D}_{\mathcal{P}}$ is composed of 2 such blocks. We use spectral normalization \cite{miyato2018cgans} in $\mathcal{D}_{\mathcal{P}}$.

\section{Experiments for single-source UDA}
Since, our proposed TriGAN has a generic framework and can handle $N$-way domain translations, we also conduct experiments for Single-Source UDA scenario where $N=2$ and the source domain is grayscale MNIST. We consider the following UDA settings with the digits dataset:

\begin{table*}[h]
    \centering
    \begin{tabular}{l|cccc}
        \hline
        \hline
         Methods & \makecell{Source\\Target}& \makecell{MNIST \\ USPS} & \makecell{MNIST \\ MNIST-M} & \makecell{MNIST \\ SVHN} \\
         \hline
         \hline
         Source Only & & 78.9 & 63.6 &  26.0 \\
         DANN~\cite{ganin2016domain} && 85.1 & 77.4 & 35.7\\
         CoGAN~\cite{liu2016coupled} &&  91.2 &  62.0 & - \\
         ADDA~\cite{Hoffman:Adda:CVPR17} && 89.4 &  - & -\\
         PixelDA~\cite{bousmalis2017unsupervised} && 95.9 & \underline{98.2} & -\\
         UNIT~\cite{liu2017unsupervised} && 95.9 &  - &  -\\
         SBADA-GAN~\cite{russo17sbadagan} && \underline{97.6} & \textbf{99.4} & \underline{61.1}\\
         GenToAdapt~\cite{sankaranarayanan2018generate} && 92.5 & - & 36.4\\
         CyCADA~\cite{hoffman2017cycada} && 94.8 & - & -\\
         I2I Adapt~\cite{murez2018image} && 92.1 &- & -\\
         TriGAN (Ours) && \textbf{98.0} & 95.7 & \textbf{66.3}\\
         \hline
         \hline
    \end{tabular}
    \vspace{2mm}
    \caption{Classification Accuracy (\%) of GAN-based methods on the Single-source UDA setting for Digits Recognition. The best number is in bold and the second best is underlined.}
    \label{tab:single_uda_sota_digits}
\end{table*}

\subsection{Datasets}

\noindent \textbf{MNIST} $\rightarrow$ \textbf{USPS}. The MNIST dataset contains grayscale images of handwritten digits 0 to 9. The pixel resolution of MNIST digits is 28 $\times$ 28. The USPS contains similar grayscale handwritten digits except the resolution is 16 $\times$ 16. We up-sample images from both domains to 32 $\times$ 32 during training. For training TriGAN 50000 MNIST and 7438 USPS samples are used. For evaluation we used 1860 test samples from USPS.\\

\noindent \textbf{MNIST} $\rightarrow$ \textbf{MNIST-M}. MNIST-M is a coloured version of grayscale MNIST digits. MNIST-M has RGB images with resolution 28 $\times$ 28. For training TriGAN all 50000 training samples from MNIST and MNIST-M are used and the dedicated 10000 MNIST-M test samples are used for evaluation. Upsampling to 32 $\times$ 32 is also done during training.\\

\noindent \textbf{MNIST} $\rightarrow$ \textbf{SVHN}. SVHN is the short form of Street View House Number and contains real world version of digits ranging from 0 to 9. The images in SVHN are RGB with pixel resolution of 32 $\times$ 32. SVHN has non-centered digits with varying colour intensities. Presence of side-digits also makes adaption to SVHN a hard task. For training TriGAN 60000 MNIST and 73257 SVHN training samples are used. During evaluation all 26032 SVHN test samples are utilized.

\subsection{Comparison with GAN-based state-of-the-art methods}
In this section we compare our proposed TriGAN with GAN-based state-of-the-art methods, both with adversarial learning based approaches and reconstruction-based approaches. Tab.~\ref{tab:single_uda_sota_digits} reports the performance of our TriGAN alongside the results obtained from the following baselines: Domain Adversarial Neural Network \cite{ganin2016domain} (\textbf{DANN}), Coupled generative adversarial networks \cite{liu2016coupled} (\textbf{CoGAN}), Adversarial discriminative domain adaptation \cite{Hoffman:Adda:CVPR17} (\textbf{ADDA}),  Pixel-level domain  adaptation \cite{bousmalis2017unsupervised} (\textbf{PixelDA}), Unsupervised image-to-image translation networks \cite{liu2017unsupervised} (\textbf{UNIT}), Symmetric bi-directional adaptive gan \cite{russo17sbadagan} (\textbf{SBADA-GAN}), Generate to adapt \cite{sankaranarayanan2018generate} (\textbf{GenToAdapt}), Cycle-consistent adversarial domain adaptation \cite{hoffman2017cycada} (\textbf{CyCADA}) and Image to image translation for domain adaptation \cite{murez2018image} (\textbf{I2I Adapt}). As can be seen from Tab.~\ref{tab:single_uda_sota_digits} TriGAN does better in two out of three adaptation settings. It is only worse in the MNIST $\rightarrow$ MNIST-M setting where it is the third best. It is to be noted that TriGAN does significantly well in MNIST $\rightarrow$ SVHN adaptation which is particularly considered as a hard setting. TriGAN is 5.2\% better than the second best method SBADA-GAN for MNIST $\rightarrow$ SVHN.